\documentclass[lettersize,journal]{IEEEtran}
\usepackage{amsmath,amsfonts}
\usepackage{algorithmic}
\usepackage{algorithm}
\usepackage{array}
\usepackage[caption=false,font=normalsize,labelfont=sf,textfont=sf]{subfig}
\usepackage{textcomp}
\usepackage{stfloats}
\usepackage{url}
\usepackage{verbatim}
\usepackage{graphicx}
\usepackage{cite}
\usepackage[utf8]{inputenc} % allow utf-8 input
\usepackage[T1]{fontenc}    % use 8-bit T1 fonts
\usepackage{url}            % simple URL typesetting
\usepackage{booktabs}       % professional-quality tables
\usepackage{amsfonts}       % blackboard math symbols
\usepackage{nicefrac}       % compact symbols for 1/2, etc.
\usepackage{microtype}      % microtypography

\usepackage{xcolor}
\usepackage{hyperref}       % hyperlinks
\definecolor{mydarkblue}{rgb}{0,0.08,0.45}
\hypersetup{colorlinks,linkcolor=red,urlcolor=mydarkblue,linktoc=page}

\usepackage{graphicx}
\usepackage{amssymb}

\usepackage{color}
\usepackage{dirtytalk}
\usepackage{comment}
\usepackage{threeparttable}
\usepackage{multirow}
\usepackage[super]{nth}
\usepackage{siunitx}
\usepackage{bm}
\usepackage[normalem]{ulem}

\newcommand{\revisedn}[1]{#1}
\newcommand{\revised}[1]{#1}
\newcommand{\removed}[1]{}

\begin{document}

%\title{GAN-Aimbots: The Next Generation of Cheating in Multiplayer Video Games}
%\title{GAN-Aimbots: Machine Learning based\\Cheating in First Person Shooters}
\title{GAN-Aimbots: Using Machine Learning for\\Cheating in First Person Shooters}

%\author{Anssi~Kanervisto,
%        Tomi~Kinnunen,~\IEEEmembership{Member,~IEEE,}
%        and~Ville~Hautam\"aki,~\IEEEmembership{Member%,~IEEE}% <-this % stops a space
% For arxiv
\author{Anssi~Kanervisto,
        Tomi~Kinnunen,
        and~Ville~Hautam\"aki% <-this % stops a space
\thanks{All authors are with School of Computing, University of Eastern Finland, Joensuu, Finland. V. Hautam\"aki is also with the Department of Electrical and Computer Engineering, National University of Singapore. E-mail: \texttt{anssk@uef.fi}, \texttt{tkinnu@uef.fi}, \texttt{villeh@uef.fi}}}
% for arxiv
%\thanks{Manuscript received December 7th, 2021; Revised March 13th, 2022; Accepted May 3th, 2022.}}

%\markboth{GAN-Aimbots}%
%{GAN-Aimbots}
% for arxiv
\markboth{}{}

% For arxiv
\IEEEpubid{}

\maketitle

\begin{abstract}
Playing games with cheaters is not fun, and in a multi-billion-dollar video game industry with hundreds of millions of players, game developers aim to improve the security and, consequently, the user experience of their games by preventing cheating. Both traditional software-based methods and statistical systems have been successful in protecting against cheating, but recent advances in the \revised{automatic} generation of content, such as images or speech, threaten the video game industry; they \revised{could be used to generate artificial gameplay indistinguishable from that of legitimate human players}. \revised{To better understand this threat}, we begin by reviewing the current state of multiplayer video game cheating, and then proceed to \revised{build} a proof-of-concept method, \textit{GAN-Aimbot}. By gathering data from various players in a first-person shooter game we show that the method improves players' performance while remaining hidden from automatic and manual protection mechanisms. By sharing this work we hope to raise awareness on this issue and encourage further research into protecting the gaming communities.
\end{abstract}

% For arxiv
%\begin{IEEEkeywords}
%video game, deep learning, cheating, security, %generative adversarial networks
%\end{IEEEkeywords}

\section{Introduction}
\label{sec:intro}
    \IEEEPARstart{V}{ideo} games attract millions of players, and the industry reports their revenue in billions of dollars. For instance, one of the biggest video game publishers, Activision Blizzard, reported more than 75 million players of their game \textit{Call of Duty: Modern Warfare (2019)} and net revenue of over 3.7 billion US dollars in the first half of 2020~\cite{blizzard2020}. The gaming communities also contain e-sport tournaments with prize pools in millions, e.g. World Electronic Sports Games 2016 contained a prize pool of 1.5 million USD. With such popularity, cheating practices similar to doping in physical sports is commonplace in video games. \revised{For example,} two public cheating communities have existed since 2000 and 2001, and have more than seven million members in total~\cite{mpgh, uc}, \revisedn{and this is only a fraction of such communities.} In these communities, users can share and develop different ways to cheat in multiplayer games. Although cheating is more difficult in in-person tournaments, cheating in online games is still prevalent. For example, UnknownCheats~\cite{uc} has 213 threads and more than a million views for the game \textit{Call of Duty: Modern Warfare (2019)}. The presence of cheaters degrades the user experience of other players, as playing with cheaters is not fun. Game developers are therefore encouraged to prevent such cheating.
    
    A common way to cheat in online games is by using tools and software (\say{hacks}) used by \say{hackers} \cite{yan2005systematic}, \revised{dating back to 1990 with the first forms of hacking with Game Genie \cite{gamegenie}}. Hacking is prohibited by game publishers, and they monitor players with anti-cheat software to detect hacking. A standard approach is to check running processes on the player's computer for known cheating software---similar to how antivirus scanners look for specific strings of binary on a machine. The publisher can also analyse players' behaviour and performance to detect suspicious activity, \revised{such as} a sudden increase in a player's performance. While hacks can be hidden from the former detection approach, the latter cannot be bypassed, as the system runs on game servers. Analysis of player behaviour is thus an attractive option for publishers.

    These data-based anti-cheats have also attracted the attention of the academic community, as well as from the game industry. Galli et al. (2011) \cite{galli2011cheating} developed hacks that provide additional information to the player and move the mouse in the game \textit{Unreal Tournament III}. The authors then used machine learning to distinguish these hackers from legitimate players. Hashen et al. (2013) \cite{alayed2013behavioral} extended this work by evaluating the accuracy of different classifiers versus different hacks. Yeung and Lui (2008) \cite{bayes_aimbot_detection} analysed players' performance and then used Bayesian inference to analyse players' accuracy and classify if a player was hacking. 
    %Our work was inspired by a commercial product in production, VACNet \cite{vacnet}, which uses neural networks to detect cheaters from player's mouse movement alone \cite{vacnet}. 
    
    All of these works have focused on \textbf{detecting} hackers with machine learning, but little attention had been given to \textbf{cheating with machine learning}. Yan and Randell \cite{yan2005systematic} have mentioned the use of machine learning for cheating, but only in the context of board games like Chess and Go and not in video games. Recent advances in machine learning allow one to generate photorealistic imagery \cite{karras2018style}, speech to attack voice biometric systems \cite{sizov2015joint} or \revised{computer agents that mimic the demonstrators~\cite{ho2016generative, fu2018learning}}. Similar methods could be applied to video games to generate human-like gameplay. \revised{We define \say{human-like gameplay} as artificially generated gameplay that is indistinguishable from genuine human gameplay, either by other humans or by automated detection systems}. If used for cheating, these methods threaten the integrity of multiplayer games, as the previously mentioned anti-cheat systems could not distinguish these cheaters.
    % Removed based on Tomi's comments
    % , similar to e.g. speech synthesis \cite{oord2016wavenet, todisco2019asvspoof}.
    \IEEEpubidadjcol
    
    To develop an understanding of this threat, we \revised{design a machine learning method for controlling the computer mouse to augment the player's gameplay and study how effective it is for cheating}. This work and its contributions can be summarised as follows:
    \begin{itemize}
        \item We review different ways people hack in multiplayer video games and methods for detecting such cheaters.
        \item We present a proof-of-concept machine learning method, GAN-Aimbot, for training a hack \revisedn{that mimics human behaviour using Generative Adversarial Networks (GANs)~\cite{goodfellow2014generative} and human gameplay data}.
        \item \revisedn{We implement GAN-Aimbot along with heuristic hacks as baseline solutions. We then collect gameplay data from multiple human participants and compare how much these hacks improve players' performance.}
        \item \revisedn{We implement an automatic cheat detection system using neural networks and use it to study how detectable the different hacks are.}
        \item \revisedn{We record video clips of human participants using these hacks and then ask another set of participants to judge if the player is cheating, based on the recording.}
        \item \revisedn{We discuss the ethical and broader impact of sharing this work.}
    \end{itemize}

    By sharing this work and associated code and data publicly, we aim to encourage further research on maintaining trust in online gaming.

\section{Cheating and anti-cheating: classic approaches} 
    
    We start with an overview of current hack and anti-cheat methods. We focus on multiplayer first-person shooter (FPS) games\removed{, which attract millions of players}. While difficult for human players to master, these games are an easy target for hackers, as quick reflexes and accurate movement are more easily implemented on a computer program than grand, strategic planning.

    % Replaced with above
    % Before moving to our work and the future vision, we must understand where we come from and where we currently stand. As our vision is on hacking that is indistinguishable from \textit{bona fide}\footnote{Genuine, legitimate, sincere. In our context ``non-cheating".} human play, we first cover real-world cases of hacks, their features and their limitations. We will also take a look at the other side of the coin on how game developers try to prevent hackers from playing and how machine learning has been utilized towards this goal. 
    %As video games run on a computer, and we as consumers have control over our computers (to a degree), a common way to achieve cheating is by hacking: using software and tools (hacks) to assist or modify the gameplay. This includes e.g. modifying the game for your advantage, using software to gain information you should not have or letting software control the game. Note that that boundary between what is counted as hacking and what is not varies. Some games allow use of so-called \textit{macros}, a predefined sequence of inputs, which allow players to e.g. fire a gun at higher fire-rates than it would normally be possible. We will refer to players using hacks as \textit{hackers}, and rest as \textit{bona fide players}\footnote{\textit{Bona fide}: Legitimate, genuine}.

    \subsection{Principles of video game hacks}
        \begin{table*}[t]
            \centering
            \begin{threeparttable}[b]
                \centering
                \renewcommand{\arraystretch}{1.5}
                \caption{Examples of hacks ``in-the-wild``, most of which are publicly and freely shared on UnknownCheats forums. These examples were picked to demonstrate different attack methods and features, prioritizing for popular games.}
                \label{tab:hacks}
                \begin{tabular}{p{0.15\textwidth}p{0.15\textwidth}p{0.1\textwidth}p{0.1\textwidth}p{0.3\textwidth}}
                \toprule
\textbf{Game}                                     & \textbf{Name}                                   & \textbf{Attack vectors}         & \textbf{Features}                 & \textbf{Attack methods}                                                                                                                                                           \\
\midrule  Call Of Duty: Modern Warfare 2                         & External ESP\tnote{1}          & External, memory (read)          & ESP, aimbot                       & Read memory for information. Draw on new window on top of game. Control mouse via OS function. Heuristic mouse movement.                                                          \\
                                                  & InvLabs M2\tnote{2}            & Internal (inject)                & ESP, aimbot, kill everyone & Read and modify game logic                                                                                                                                                        \\
\hline ARMA 2 (DayZ mod)                & DayZ Navigator\tnote{3}        & External, memory (read)          & ESP                               & Read memory for information.  Show enemy locations on browser-based map                                                                       \\
                                                  & DayZ "Wallhack"\tnote{9}       & External, modify files & ESP, shoot through walls          & Remove objects from game by removing files.                                                                                            \\
                                                  & DayZ Control Players\tnote{8}  & Internal (inject), exploit       & Misc.                             & Take control of characters of other players using game's functions.                                                                                                               \\
\midrule DayZ                                              & DayZ Ghost\tnote{7}            & External, memory (read/write)    & Fly-hack                          & Write desired location of player into memory faster than game would modify it                                                                                                     \\
\midrule Counter Strike: Global Offensive & Charlatano\tnote{4}           & External, memory (read)          & ESP, aimbot                       & Read memory for information. Control mouse via OS function. Advanced heuristics for human-like behaviour (short movement, Bézier curves). \\
                                                  & Neural Network Aimbot\tnote{5} & Displayed image                  & Aimbot                            & Use CV to detect enemies on the screen. Control mouse via OS function. Heuristic mouse movement.                                                                                      \\
\midrule PLAYERUNKNOWN'S BATTLEGROUNDS                     & PUBG-Radar\tnote{6}            & Network (read)                   & ESP                               & Read network packets for information. Draw info on a separate window.                                                                                                             \\
\midrule Battlefield 4                                     & BF4 Wallhack\tnote{10}         & Unknown, exploit                 & ESP, shoot through walls          & Possibly exploiting the fact server does not check hits against walls.                \\
\bottomrule
\end{tabular}
                \begin{tablenotes}
                    \item [1] \url{https://www.unknowncheats.me/forum/call-duty-6-modern-warfare-2/64376-external-boxesp-v5-1-bigradar-1-2-208-a.html}
                    \item [2] \url{https://www.unknowncheats.me/forum/call-of-duty-6-modern-warfare-2-a/91102-newtechnology_publicbot-v4-0-invlabs-m2.html}
                    \item [3] \url{https://www.unknowncheats.me/forum/arma-2-a/80397-dayz-navigator-v3-7-a.html}
                    \item [4] \url{https://github.com/Jire/Charlatano}
                    \item [5] \url{https://www.unknowncheats.me/forum/cs-go-releases/285847-hayha-neural-network-aimbot.html}
                    \item [6] \url{https://www.unknowncheats.me/forum/pubg-releases/262741-radar-pubg.html}
                    \item [7] \url{https://www.unknowncheats.me/forum/dayz-sa/228162-ghost-mode-free-camera-tutorial-source-video-compatible-online-offline.html}
                    \item [8] \url{https://www.unknowncheats.me/forum/arma-2-a/114967-taking-control-players.html}
                    \item [9] \url{https://www.unknowncheats.me/forum/arma-2-a/77733-dayz-wallhack-sorta.html}
                    \item [10] \url{https://www.youtube.com/watch?v=yF7CyqbSS48}
                \end{tablenotes}
            \end{threeparttable}
        \end{table*}
        
\begin{figure}[t]
            \centering
            \includegraphics[width=1.0\columnwidth]{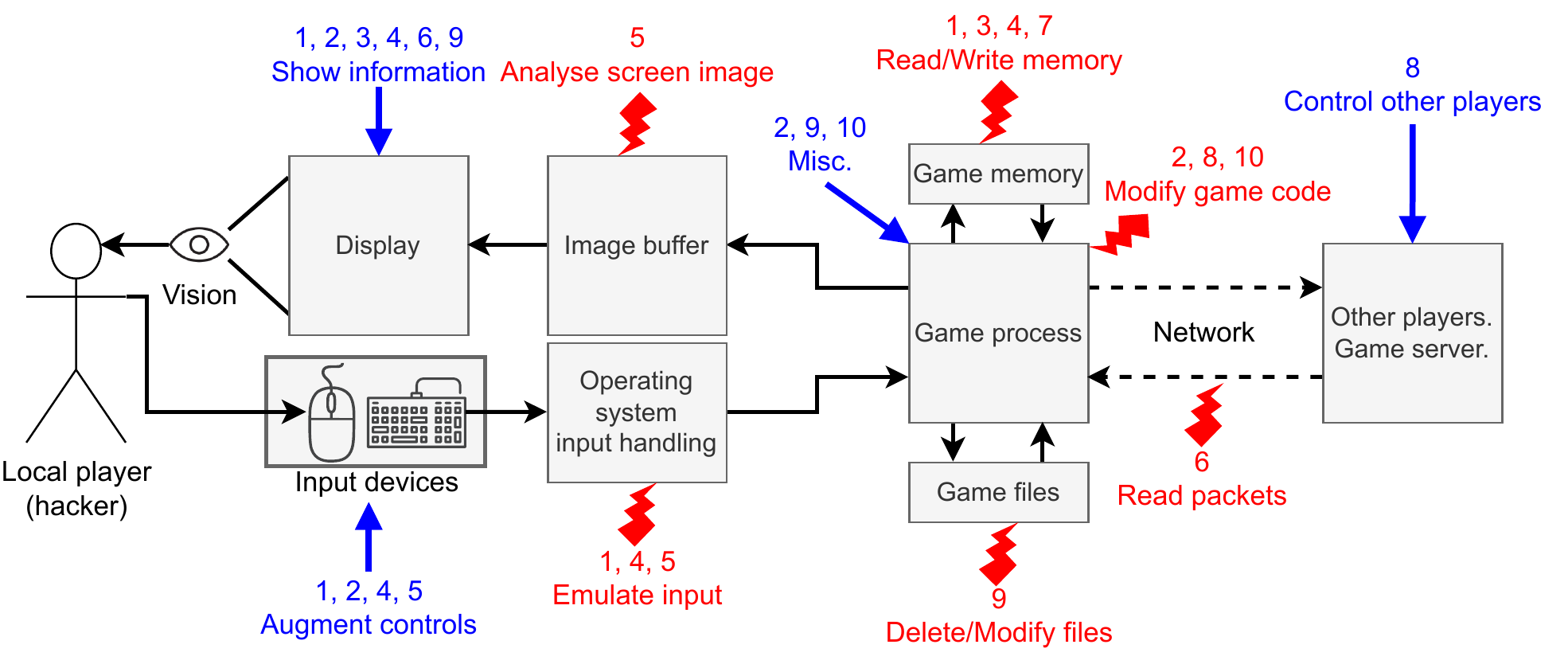}
            \caption{Different attack vectors used by hacks which are listed in Table \ref{tab:hacks} (see the list under Table \ref{tab:hacks}). Red, lightning symbols indicate how hacks attack the information flow (how hacks work), while blue arrows show what hacks add to a hacker's gameplay (what hacks do).}
            \label{fig:attacks}
        \end{figure}

        Much like attacking any system with the intent of modifying it, hacking involves knowing how a game works and then using this knowledge to achieve what the hacker wants. We draw inspiration from biometric security \cite{ratha2001enhancing, roberts2007biometric} and present different attack vectors in Figure \ref{fig:attacks}. An attack vector is a point in information flow that the hacker can access to implement their hack. \removed{Identifying these points requires an understanding of how the game code functions.} One must reverse engineer the game logic \removed{or memory structures} before a game can be attacked. For example, if the hacker can find the function that reduces their player's health when they take damage, they can modify this function to not subtract health, which renders them invulnerable. Similar modifications can be made to game files or network packets. Reverse engineering is arguably the hardest part of creating hacks and the most prominent topic in the game hacking communities like UnknownCheats~\cite{uc}.
        %\todo{Following text discusses hacks in Table 1 (which contains references/links), but how to refer to them here?}
        
        While the above attack vectors mainly affect personal computers (PCs), for which the hacker has access to the whole computer system, closed systems like game consoles (e.g., Sony's PlayStation and Microsoft's Xbox) \revisedn{can also be modified to gain full access to the system (\say{jailbreaking})}. \revised{Game streaming services, like Google Stadia, Nvidia GeForce Now and Sony PlayStation Now, further restrict this access by streaming the screen image and audio over a network, rendering system access impossible.} However, two attack points remain; the user output (display and audio) and the user input devices (mouse and keyboard). While these are not trivial to attack, we believe these elements will be vulnerable to attacks in the future, which we discuss in Section \ref{sec:data-based-cheating}. For now, we focus on hacking done on PCs.
        
        % The most common platform for games of interest \needcite, personal computers, have many such points as players have full access to the system. Game consoles like Sony PlayStation or Microsoft XBox make it harder to access the software running on machine, but can still be \say{rooted} for full access, similar to a personal computer. Game streaming services (Google Stadia, NVIDIA GeForce Now, Sony PlayStation Now) take this restriction to a limit, where game runs on a remote server and only the video and audio of the game is sent over the network to the player. This limits attack-surfaces to the image/audio given to player and to the controls they use, making hacking harder but not impossible, to which we return later. 
        
        %\textit{Game-streaming services} work by running the program on a remote server, sending the rendered images to the player's device and then receiving player's input. This limits the number of attack-points considerably, leaving the rendered image as the only possible source of information, and emulation of user input as an attack. If this is combined with a locked down system, where the player has no other access to system than the provided input devices and a screen, the only way to cheat would be to use cameras to record the screen and use robotic apparatus to control input devices. \todo{move to something more sensible}
    
    \subsection{Hacking in the wild}
    \label{sec:hacks-in-the-wild}
        
        Table \ref{tab:hacks} includes a set of examples of publicly available hacks for popular PC games. They are often categorised by their \say{features} (advantages they provide to the player) and how they are implemented. Internal hacks modify the program code via injection of the hacker's code. This technique remains a classic and active form of hacking to this day but is easily detectable due to its intrusive nature. External hacks do not modify the game code. Rather, they rely on reading and writing game memory, inspecting network traffic or modifying game files to achieve desired effects. Computer vision (CV) hacks neither read nor write any game-related information Instead, they capture the image shown to the player and extract necessary information via CV techniques. \revised{These include reading the colour of a specific pixel, shape detection or object detection  \cite{redmon2016you}.}
        
        \revised{The most common features are \say{wall hacks} and \say{radar hacks}, which show the location of objects of interest, such as enemies.} We group these types of hacks using a commonly used term, extrasensory perception (ESP). ESP encompasses hacks that provide information not normally available to players. Another common feature, albeit one specific to FPS games, is \say{aimbots}. These help players aim at enemies, an act that normally requires accurate reflexes and is a core part of the competitive nature of shooter games (e.g., Quake, Unreal Tournament, Call of Duty, Battlefield). ESP and aimbots are the most common features across all FPS games, as they \revised{provide a strong advantage} and can be implemented via most of the attacks listed above. For example, to implement an ESP feature, the hacker only needs information about the locations of the players and which of the players is the hacker's player. With this information alone, the hacker can draw a 2D map that indicates the locations of other players relative to theirs. Aimbots are similarly easy to implement, and only require the current aiming direction of the player. The hacker can then calculate how the aim should be adjusted towards a nearby enemy. 
        
        While common, ESP and aimbots are not the only possible features. By exploiting game-specific mechanics, other hacks can remove game objects by removing files (e.g., \say{DayZ Wallhack}), update player location more frequently than the game does to allow flight (e.g., \say{DayZ Ghost}) or shoot through walls (e.g., \say{BF4 Wallhack}). These are only limited examples of different hacks people have devised, but they generally are easily noticeable by other players or can be prevented by developers.

        An open-source hack more similar to our core topic of human-like behaviour is \say{Charlatano}, which advertises itself as a \say{stream-proof} hack. This hack would not be detected by people observing the gameplay of the hacker and is achieved via a human-like aimbot, which adjusts aim in natural and small increments to avoid suspicion. These features are based on heuristics and on visual observations of what looks human-like to humans. Another hack, the \say{Neural Network Aimbot}, uses the deep neural network \say{YOLO}~\cite{redmon2016you} to detect targets visible on the screen and then directs aim towards them by emulating mouse movement. The screen image is captured in a manner that mimics game-recording software, and the mouse is moved with operating-system functions, rendering this hack virtually invisible to game software. If combined with an external video-capture device and an independent micro-controller emulating a mouse, this hack would be \removed{completely} independent of the machine that runs the game. The behaviour of the player, and specifically the effect of aimbot on how a player moves their mouse, would be the only possible source for detecting this hack.
    
    \subsection{Countermeasures against hacking}
    \label{sec:anti-cheats}
    
        \begin{table*}[t]
            \centering
            \begin{threeparttable}[b]
                \centering
                \renewcommand{\arraystretch}{1.2}
                \caption{List of common anti-cheat software found in games, along with the description of methods they use for detecting hackers.}
                \label{tab:anti-cheats}
                \begin{tabular}{lll}
\toprule
\multicolumn{1}{l}{\textbf{Name}} & \multicolumn{1}{l}{\textbf{Type}} & \multicolumn{1}{l}{\textbf{Features}}                                                     \\ \midrule
BattleEye \cite{battleye}                         & Traditional                       & Signature checks, screenshots                                                            \\
Punkbuster \cite{punkbuster}                        & Traditional                       & Signature and driver checks, screenshots                                                 \\
FairFight \cite{fairfight}                         & Statistical                       & Various. Track player statistics, track if player aims through walls.                    \\
EasyAntiCheat \cite{eac}                     & Traditional + Statistical         & Signature checks, low-level access to memory, track player statistics.                   \\
Valve Anti-Cheat \cite{vacnet}                  & Traditional + Statistical         & Signature checking, deep learning-based aimbot detector based on mouse movement (VACNet) \\
\bottomrule
\end{tabular}
            \end{threeparttable}
        \end{table*}
        
        To prevent hacking and cheating, passive methods include designing game flow such that hacking is not possible or obfuscating the compiled game binary such that reverse engineering becomes more difficult. An example of the former option is the game \say{World of Tanks}, in which the player client does not receive any information about enemy locations until they are within line of sight, rendering \revised{ESP hacks difficult to implement}. The latter option can be a by-product of a digital rights management (DRM) system or an anti-tamper system like that of Denuvo \cite{denuvo}, which encrypts the executable code and prevents the use of debugging tools for analysing game flow. 
        
        If the creation or use of hacks was difficult or required expensive hardware to execute, hacking would be limited, much like how long passwords are considered secure because randomly guessing them is virtually impossible. Unfortunately, as game logic and code are the same for all players a game, only one person needs to create and share a hack before everyone else can use it, much like the unauthorised sharing of game copies. As such, passive methods alone do not reliably prevent hacks unless they are guaranteed to prevent hacking. Active prevention methods are thus necessary.
        
        Active prevention methods aim to detect players using hacks and then remove them from the game, temporarily or permanently (by banning the player). A simple method is to use human spectators who look for suspicious players, but this option does not scale with high numbers of players. Human spectators are also susceptible to false alarms; genuinely proficient players may be flagged as hackers.\footnote{For an example of such players, search for gameplay footage of professional FPS players. Here is an example of a professional player in Battlefield 4: \url{https://www.youtube.com/watch?v=6-Kt2duNKtY}.} CS:GO Overwatch~\cite{csoverwatch} crowd-sources this by allowing the game community to judge recordings of other players. 
        
        A more common approach is to use anti-cheat software to identify these hackers. Table \ref{tab:anti-cheats} lists common anti-cheats found in modern games. A traditional approach is to search for specific strings of bytes in computer memory \removed{(i.e., signatures)} obtained by analysing publicly available hacks. The finding of such a signature indicates that a known hack is running on the machine, and the player is flagged for hacking. This method alone is effective against publicly shared hacks but is not helpful against private \removed{(non-shared)} hacks. 
        % Commented out as not relevant
        %Another method is to take occasional screenshots of the game and scan them for drawings that are not supposed to be there. This detects internal hacks that use the game functions to draw new information on the screen, which are then visible in these screenshots and easy to detect by e.g. checking for sharp changes in the colour of the pixels. 
        
        As with any attack-defend situation, with hackers being the attackers and anti-cheats the defenders, there is a constant battle between building better defences and avoiding these defences with better attacks. Signature checks can be avoided by \removed{randomising and} obfuscating the binary code of hacks.
        % Commented out as talk about screenshots was also removed above
        %and as external hacks may draw their information on a separate window from that of the game, their drawings are not visible in the screenshots taken by anti-cheats.
        In turn, anti-cheats can search for suspicious memory accesses to the game process. Hacks can hide this evidence by using lower levels in the security rings (\say{kernel-level}), which cannot be scanned by anti-cheats running on the higher levels. Naturally, anti-cheats can adapt by installing themselves on this lower level. 
        % Commented out as bit of a meander
        %Naturally, anti-cheats can be installed on such low-levels as well but this raises questions about privacy and security as such a software could theoretically probe other memory regions and read sensitive information.
        However, as PCs are ultimately owned by their users, a hacker could modify their system to fool anti-cheats. Luckily for defenders, studying a user's machine is not the only way to detect hackers.

    \subsection{Data-based approach to detecting hackers}
        Anti-cheats like FairFight \cite{fairfight} and Valve Anti-Cheat (VAC) \cite{vacnet} detect hackers by their performance and behaviour, in addition to \removed{any possible} traditional techniques. Hackers can be detected using heuristics (e.g., too many kills in a short time is suspicious) or with statistical models derived from data (e.g., VACNet \cite{vacnet} and related work, \cite{galli2011cheating, bayes_aimbot_detection}). As these types of anti-cheats are executed on the game servers and use the same data the player client must send to play the game \revised{in multiplayer games}, they cannot be bypassed by hackers like traditional anti-cheats can be. Unfortunately, these methods do have their weakness: errors in classification \cite{duda2012pattern-chapter2, brummer2010measuring}.
        
        Identifying hackers is a binary-classification task: is the given player a hacker or \textit{bona fide}?\footnote{Genuine, legitimate, sincere. In our context ``non-cheating".} With two possible cases and two possible decisions, there are four outcomes, two of which are desired (correct decisions) and two which are errors. In a false positive (FP) case, a \textit{bona fide} player is \revised{predicted as a} hacker, and in a false negative (FN) case a hacking player is \revised{predicted as a} \textit{bona fide} player. Since the two classes are unlikely to be perfectly distinguishable (i.e., mistakes will inevitably occur), \revised{system designers must balance between user convenience and security}.
        %should the system be more aggressive to flag players as hackers (potentially leading to more false positives), or be very careful in doing so (potentially leading to more false negatives)? 
        A game developer likely wants to develop a system that exerts caution when flagging people hackers, as \revised{FPs harm \textit{bona fide} players, which in turn harm the game's reputation.} However, if an anti-cheat only analyses player performance, then a hacker can adjust their hacks to mimic top-performing, \textit{bona fide} players. These hackers are still better than most players and they avoid detection by an anti-cheat system that has been tuned not to flag top-performing players. This approach clearly has limitations, which brings us to the analysis of player behaviour.
        
        For example, FairFight gathers information on how a player moves and aims at targets through walls~\cite{ffavoidbans}. If a player often aims at enemies behind obstacles and walls, this behaviour signals that a hack is being used, and the player can be flagged. VACNet employs a similar approach\removed{but is fully automated}: a deep learning model has been trained to classify between \textit{bona fide} and hacking players according to their mouse movement, using a dataset of known hackers and \textit{bona fide} players \cite{vacnet}. This kind of anti-cheat does not rely on high performance to detect a hacker, as it attacks the very core of hacking: altering the player's behaviour to improve performance. A hack cannot improve a player's performance if it does not alter their behaviour. 
        %Hacks like Charlatano claim to be \say{stream proof}, meaning it is not detectable by human observers, by only changing the behaviour a small amount, but there is no solid evidence of this.
        
        % Removed for continuation sake to the next section
        %Some of the hacks mentioned above aim to avoid detection by these with slower aimbots which do not appear too suspicious, or with complex human-like mouse movement curves like in the case of Charlatano. However it is not certain how effective these methods are, apart from the bold claims on their website. The best public knowledge we have comes from the previous academic work \cite{galli2011cheating, bayes_aimbot_detection}, but they share a shortcoming: the aimbots they used for their experiments were crude, where the aimbot moves mouse directly on top of the target in one step or in very fast and accurate motion, both of which are easily detectable by human players (see Section \ref{sec:human-evaluation-results} for our results on this). As it stands, we do not know how well the anti-cheat systems detect state-of-the-art hacks. Worse yet, given that  

\section{Data-based approaches to hacking}
    \label{sec:data-based-cheating}

    % Commented out to save space
    %\begin{figure}[t]
    %    \centering
    %    \includegraphics[width=.5\columnwidth]{material/ml-aimbots-cloning.pdf}
    %    \caption{ \todo{This could be made much smaller, change innocent to \textit{bona fide}} A hypothetical situation. A malicious player clones the behaviour of a \textit{bona fide} player, uses it to play game and thus gains an unfair advantage. The anti-cheat is designed to pass innocent players and ban hacking players, but what can it do to the cloned player? If the cloning is perfect, then anti-cheat must pass the cloning player, otherwise it would also ban the \textit{bona fide} player.}
    %    \label{fig:cloning-attack}
    %\end{figure}
        
    What if the behaviour of a hacker is indistinguishable from a \textit{bona fide} human player? Consider a situation in which a hacker can perfectly clone the gameplay behaviour of a professional player using advanced methods. If the professional player is not flagged for cheating, then this hacker would not be flagged either, but clearly, the latter party is cheating. 
    
    \removed{This situation has inspired our vision for the future of video games and hacking.} For this paper, the scope of this cloning is limited to mouse movement. If this mimicry is combined with hardware that emulates mouse control and captures a screen image, and a neural network aimbot described earlier, this hack could not be detected by any of the anti-cheats methods described above.

    While this example is extreme, recent deep learning advances enabled the generation of various realistic content that is indistinguishable from natural content, like images of faces~\cite{karras2018style} and speech~\cite{oord2016wavenet}. Given these results, cloning human behaviour in a video game does not seem unrealistic. \revised{To assess the severity of this risk}, we design a proof-of-concept method \revised{in the context of aimbots}, dubbed GAN-Aimbot, and assess both its effectiveness as an aimbot and its undetectability. \revised{We limit the scope to aimbots to focus the scope of the experiments, and also because an undetectable aimbot would disrupt FPS games.} To the best of our knowledge, we are the first to openly share such a method. Although our method could be used for malicious purposes, we believe it is better to openly share this information to spur discussion and future work to prevent this type of cheating. We further discuss the positive and negative impact of sharing this work in Section \ref{sec:impact}.

    %The largest practical issue in creating this hack is the need for a dataset of human play, preferably from various players, to capture the nature of \say{human-like} mouse movement. However the work by Aytar et al. (2018) demonstrate how different video recordings of humans playing games can be used to train reinforcement learning agents \cite{aytar2018playing}, and similar techniques could be applied here to use videos of human players to learn their mouse movement.
    
\section{GAN-Aimbot: human-like aimbot with generative adversarial networks}
    \label{sec:gan-aimbot}
    \begin{figure}[t]
        \begin{center}
        \centerline{\includegraphics[width=1.0\columnwidth]{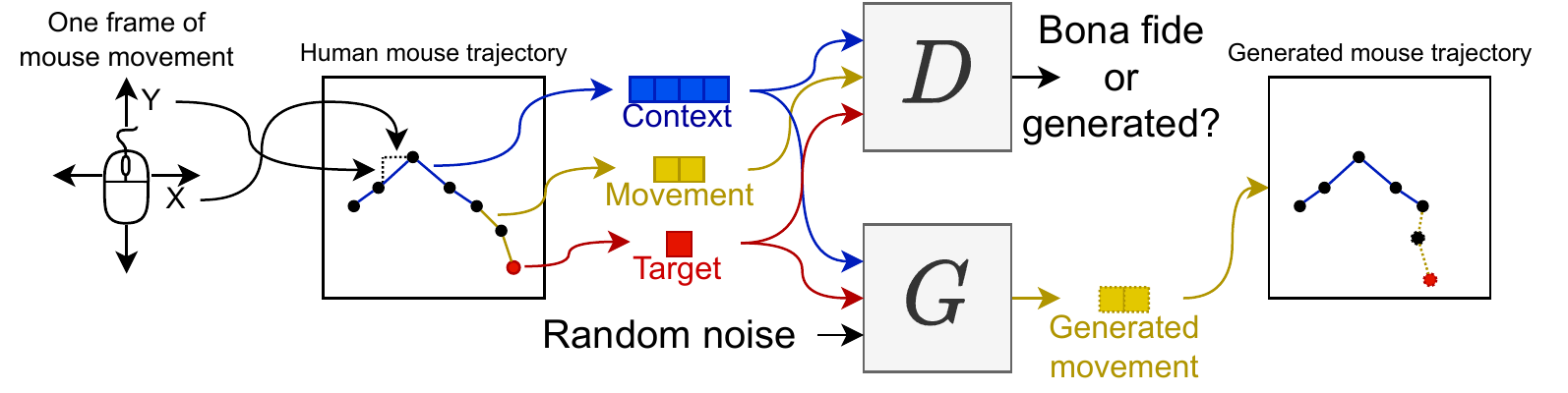}}
        \caption{An illustrative figure of the discriminator $D$ and generator $G$ networks and their inputs/outputs, with a context size of four and two movement steps. Note that ``target" is an absolute location with respect to where the trajectory begins, while other values are one-step changes in location. This setup corresponds to training (Algorithm \ref{algo:gan}), where ``target" is taken from human data.
        %Only the generator part is needed during deployment as an aimbot, where ``target" can be set to desired point on screen, e.g. enemy character.}
        }
        \label{fig:gan}
        \end{center}
    \end{figure}
    
    \begin{algorithm}[t]
    \caption{Pseudo-code for training the GAN-Aimbot.}
    \label{algo:gan}
    \begin{algorithmic}
        \STATE \textbf{Input:} Dataset containing human mouse movement $\mathcal D$, generator $G$ and its parameters $\bm \theta_G$, discriminator $D$ and its parameters $\bm \theta_D$, number of discriminator updates $N_D$, Adam optimizer parameters $\text{Adam}_D$ and $\text{Adam}_G$ and weight clip parameter $w_\text{max}$.
        \WHILE {training}
            \FOR {$n \in \{1 \ldots N_D\}$}
                \STATE //Sample human data 
                \STATE $\bm x_D, \bm y_D \sim \mathcal D$ 
                \STATE //Generate steps 
                \STATE $\bm z \sim \mathcal N(\bm 0, \bm I)$ 
                \STATE $\bm g = G(\bm z | \bm y_D)$ 
                \STATE //Update discriminator 
                \STATE $\mathcal L_D = D(\bm x_D | \bm y_D) - D(\bm g | \bm y_D)$ 
                \STATE $\bm \theta_D = \bm \theta_D - \text{Adam}_D(\nabla_{\bm \theta_D} \mathcal L_D)$ 
                \STATE //Clip weights 
                \STATE $\bm \theta_D = \text{clip}(\bm \theta_D, -w_\text{max}, w_\text{max})$
            \ENDFOR
            \STATE //Sample conditions 
            \STATE $\bm y_G \sim \mathcal D$ 
            \STATE //Generate steps 
            \STATE $\bm z \sim \mathcal N(\bm 0, \bm I)$ 
            \STATE $\bm g = G(\bm z | \bm y_G)$ 
            \STATE //Create generator loss + target point loss \eqref{eq:dist}
            \STATE $\mathcal L_G = D(\bm g | \bm y_g) + \text{dist}(\bm g, \bm t)$ 
            \STATE //Update generator 
            \STATE $\bm \theta_G = \bm \theta_G - \text{Adam}_G(\nabla_{\bm \theta_G} \mathcal L_G)$ 
        \ENDWHILE
    \end{algorithmic}
    \end{algorithm}

    The task of an aimbot is to move the aim-point (where the weapon points at, illustrated by the green cross-hair in Figure \ref{fig:vizdoom_env}) to the desired target (e.g., an enemy player) by providing \removed{horizontal and vertical} mouse movement, which would normally be provided by the human player. In each game frame, the aimbot moves the mouse $\Delta x$ units horizontally (to change the yaw of the aim direction) and $\Delta y$ units vertically (to change pitch).
    
    \subsection{Generative adversarial networks \revised{for aimbots}}
        \removed{If hackers adopt the machine learning methods already employed by anti-cheat systems, the aim of the two parties is quite clear: hackers try to train a model that creates human-like behaviour, while an anti-cheat tries to discriminate between hackers from \textit{bona fide} players. This interaction between anti-cheat and hacker is comparable to the training of generative adversarial networks (GANs) \cite{goodfellow2014generative}.}
        \revised{GAN-Aimbot, as the name suggests, is built on GANs~\cite{goodfellow2014generative}, which consists of two components.} A generator $G$ tries to create content that looks as realistic as possible (i.e. similar to the training data), while a discriminator $D$ tries to distinguish between data from the training set and data from the generator. \removed{With enough training, GANs and their derivatives have been used to imitate expert's behaviour in control \cite{ho2016generative} or generate human-like speech \cite{binkowski2020high}.} In GAN-Aimbot, the generator generates human-like mouse movement and a discriminator tries to distinguish this movement from the recorded, \textit{bona fide} human data. After training, we can use the generator as an aimbot in the game. If the generator is able to fool the discriminator at the end of the training, we assume the same applies to other anti-cheat systems, similar to how computer-generated imagery can fool human observers by appearing natural~\cite{rossler2019faceforensics++}. 
    
        Training the hack to generate human-like mouse movement is only one part of this task. The hack should also redirect aim towards potential targets. \removed{This process calls for conditional generation, where the given target is the condition.} \revised{We use a conditional GAN \cite{mirza2014conditional}, where the desired target is given as a condition to both parts.} We also provide mouse movement from previous frames in the condition (context), which allows the generator to adjust generation according to previous steps and the discriminator to spot any sudden changes.
        
        During training, we use the aim-point of the \textit{bona fide} data a few steps after the context as a target. The generator is then tasked with generating these steps (movement) such that they appear \textit{bona fide} and such that the final aim-point is close to the target. We use several steps instead of just one to provide more generated data for the discriminator to classify, and we also allow the generator to generate different steps to reach the goal. See Figure \ref{fig:gan} for an illustration.

        \revised{This method is similar to imitation learning methods using adversarial training setup~\cite{ho2016generative, fu2018learning}, where the discriminator attempts to distinguish between the computer agent being trained and the demonstration data. The discriminator is then used to improve the agent. However, we have notable differences in our proposed setup:
        \begin{enumerate}
            \item We train the system much like the original GANs~\cite{goodfellow2014generative} on a fixed dataset (we do not need the game as a part of the training process) and use the generator directly as a decision-making component.
            \item The generator is used in conjunction with a human player and not as an autonomous agent.
            \item We augment the discriminator training loss with an additional task (move aim point closer to the target).
            \item We explicitly study both the behavioural performance (akin to imitation learning performance) and distinguishability (akin to GAN literature) from demonstration data of the generator, not just one.
        \end{enumerate}
        }
    
        \revisedn{The \say{Neural Network Aimbot} hack described in Section~\ref{sec:hacks-in-the-wild} also uses neural networks, but only to detect targets using image analysis; mouse movement is still rule-based. As such, GAN-Aimbot and Neural Network Aimbot are complementary approaches. In this work, we can obtain the location of targets directly from the game environment, and as such, we do not include Neural Network Aimbot as part of our experiments.}
    
    \subsection{Training GAN-Aimbot}
        Let context size be $c$ frames and the number of generated steps $g$ leading to each data point sampled from the \textit{bona fide} dataset be length $d := c + g$ frames. Each sample is a trajectory of mouse movements $(\Delta x_1, \ldots, \Delta x_{d})$ and $(\Delta y_1, \ldots, \Delta y_{d})$, where $\Delta$ is the amount of mouse movement in one frame. The last $g$ frames are used to create the target $\bm t := (\sum^{d}_{i=c} \Delta x_i, \sum^{d}_{i=c} \Delta y_i)$, which represents where the aim-point aims $g$ frames after the context. The condition then contains the context and target $\bm i := (\Delta x_1, \ldots, \Delta x_{c}, \Delta y_1, \ldots, \Delta y_{c}, t_1, t_2)$. The discriminator accepts this context as input as well as $g$ steps $\bm o := (\Delta x_{c + 1}, \Delta x_{c + 2}, \ldots, \Delta x_{d}, \Delta y_{c + 1}, \Delta y_{c + 2}, \ldots, \Delta y_{d})$, which represent mouse movement $g$ frames after the context. These steps $\bm o$ can be from the \textit{bona fide} dataset or generated by the generator. Finally, the generator accepts as input a Gaussian-noise vector $\bm z := \mathcal N(\bm 0, \bm I), \; \bm z \in \mathbb R^{k}$ and condition $\bm i$ to generate steps $\bm o$. We use $k = 16$ which we found to be sufficient to generate good results. For a concrete example, the blue lines in Figure \ref{fig:gan} correspond to context with $c=4$, the yellow lines to generated steps with $g = 2$ and the red point is the target.
        
        The training procedure is presented in Algorithm \ref{algo:gan}. The procedure uses methods from Wasserstein GAN \cite{arjovsky2017wasserstein}, which stabilise training over standard GAN. \revised{Contrary to the GAN literature, the discriminator output is maximized for the generated content. This is to stay consistent with the classification labels, where positive labels mean a positive match for a hacker.} \removed{The discriminator is trained to maximise the output when input is from the generator and minimise the output when input is from the \textit{bona fide} data.\footnote{Note that this approach is contrary to other GAN literature. To stay consistent, larger output indicates a higher likelihood of a hacker.} The generator is trained to minimise the output of discriminator on its generated samples.}For a set of randomly sampled \textit{bona fide} conditionals $\bm i_G, \bm i_D$ (which include targets $\bm t_G, \bm t_D$), corresponding \textit{bona fide} mouse movement $\bm o_D$ and random Gaussian vectors $\bm z_G, \bm z_D$, we compute two loss values
        \begin{align}
            \mathcal L_G &:= D(G(\bm z_G | \bm i_G)) + \text{dist}(G(\bm z_G | \bm i_G), t_G)
            \label{eq:loss_generator}\\
            \mathcal L_D &:= D(\bm o_D | \bm i_D) - D(G(\bm z_D | \bm i_D) | \bm i_D) 
            \label{eq:loss_discriminator},
        \end{align}
        where $\text{dist}(\cdot, \cdot)$ calculates the Euclidean distance between the final aim-point and the target
        \begin{align}
            \bm g &:= G(\bm z_G | \bm i_G) = (\Delta x_1, \ldots, \Delta x_{g}, \Delta y_1, \ldots, \Delta y_{g}) \\
            \text{dist}(\bm g, \bm t) &:= \sqrt{ \left ( \left [ \sum^g_{i=1} g_i \right ] - t_0 \right )^2 + \left ( \left [ \sum^{2g}_{i=g} g_i \right ] - t_1 \right )^2}
            \label{eq:dist}.
        \end{align}
        Afterwards, networks are updated to minimise their corresponding losses using stochastic gradient descent and Adam optimisation \cite{adam}.\removed{which adapts gradient step sizes according to previous steps.} Discriminators' weights are clipped to preserve the smoothness of the output \cite{arjovsky2017wasserstein}.
        
        After training, we use the generator to create the proposed aimbot. We present the generator with a target point (the closest enemy to the aim-point), context and a fixed random vector; generate the next $\bm o$ mouse movements for $g$ frames; and use the output for the first frame to move the player's aim-point. This allows the aimbot to correct mouse movement in each game frame. We fix the random vector in one game to keep the generated behaviour consistent throughout the game's duration.
        
        % Removed as not really relevant
        % It should be noted that this selection of conditional- and Wasserstein GANs and hyperparameters (Section \ref{sec:training-gan}) was a result of manual trial-and-error, along with the hyperparameters discussed later. While these may be specific to this environment, it only takes a single hack developer to adjust these settings to work in other games, after which they can share the hack with others who do not need knowledge of GANs.

\section{\revisedn{Overall} experimental setup}
    \begin{figure}[t]
        \begin{center}
        \centerline{\includegraphics[width=0.75\columnwidth]{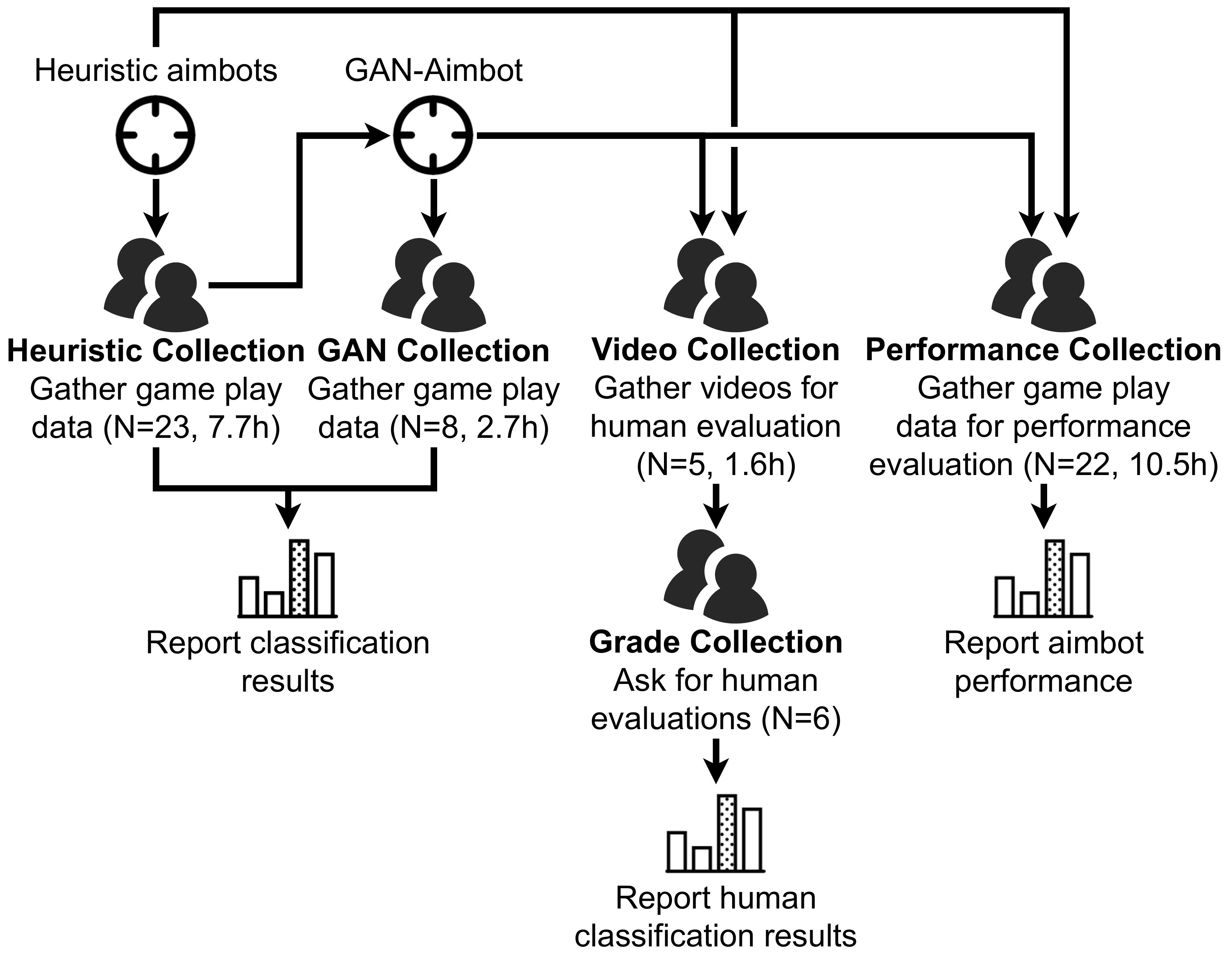}}
        \caption{\revisedn{An overview of the experiments and data collection. Data collection phases indicate the number of participants and the total amount of game time recorded.}}
        \label{fig:study-flow}
        \end{center}
    \end{figure}

    \revised{For assessment we use two research questions (RQs)}:
    
    \begin{enumerate}
        \item[\textbf{RQ1}] Does the GAN-Aimbot improve players' performance?
        \item[\textbf{RQ2}] Does the GAN-Aimbot remain undetectable by human observers and automatic anti-cheat systems?
    \end{enumerate}
    
    We answer these questions via an empirical study where we create a game environment; collect gameplay of human participants playing with and without aimbots; and assess how distinguishable gameplay with aimbots is from \textit{bona fide} human play, according to both an automated anti-cheat system and human observers. We compare the results to those of heuristic aimbots which use techniques outlined in Table \ref{tab:hacks}. The experiment process is outlined in Figure \ref{fig:study-flow}. \revisedn{The environment and aimbots are described in this section. The following sections describe individual experiments, including their setup and results.}
    
    The experiment code and collected data are available at \url{https://github.com/Miffyli/gan-aimbots}.

    \subsection{Game environment}
        \label{sec:env}
        
        \begin{figure}[t]
            \begin{center}
            \centerline{\includegraphics[width=0.7\columnwidth]{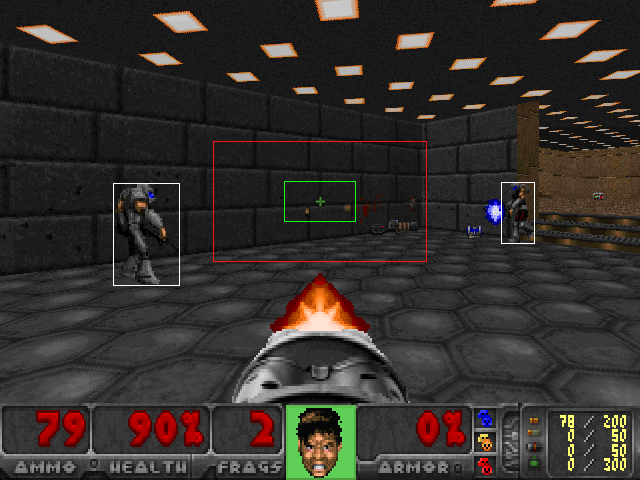}}
            \caption{ViZDoom environment used for the experiments. The player can also aim vertically in this implementation of Doom. Red and green boxes represent field-of-views of the aimbots, red for \textit{strong} and GAN-Aimbot and green for \textit{light} aimbot. Aimbots activate automatically when a target (white boxes) is inside the field-of-view. The white, green and red boxes are not visible to the player.}
            \label{fig:vizdoom_env}
            \end{center}
        \end{figure}
        
        We use Doom (1993) via ViZDoom library \cite{vizdoom} as a game, as depicted in Figure \ref{fig:vizdoom_env}. Doom is one of the first FPS games. While dated, Doom still includes core mechanics of FPS games, namely navigating the map to find enemies and killing them by aiming and firing weapons. We chose this game and library as ViZDoom allows granular control over game flow while remaining lightweight, which allows us to distribute the data collection programs to participants over the Internet\removed{without installation requirements}.
        
        The human players play games of \say{deathmatch} on a single map, where players have to score the highest number of kills to win. \revised{Everyone plays against each other}. A single human player plays with six computer-controlled players of varying strength; some exhibit higher-than-normal mobility and speed, making them difficult targets to hit. Players can find weapons stronger than the starting weapon (a pistol), but most weapons require multiple hits before they kill a target, highlighting the need for accurate aiming. So-called \say{instant-hit} weapons are of the most interest to our work---the projectiles of these weapons hit the target location immediately. These weapons reward the accurate aiming of the player and benefit the most from \revised{the use of} aimbots.
        
        \revisedn{To collect data from participants, we provided them with an executable binary file that provided information about the study, asked for consent, provided instructions and finally launched the game with the desired settings. The software recorded the participants' actions, mouse movement per frame (in degrees), damage done, weapons used, current ammunition, number of kills and deaths, and the bounding boxes of the enemies on the screen in each frame of the game ($35$ frames per second). Participants were recruited amongst lab members, authors' colleagues and first-person shooter communities to include data from players with varying skill levels.}
        
    \subsection{Heuristic aimbots}
        \begin{table}[t]
            \centering
            \caption{Settings of the different aimbots}
            \label{tab:aimbot-settings}
            \begin{tabular}{lll}
            \toprule
            \textbf{Aimbot} & \textbf{Activation range} & \textbf{Movement per axis}  \\ \midrule
            Light  & 5 degrees        & $0.4 \times \text{distance to the target}$     \\
            Strong & 15 degrees       & $0.6 \times \text{distance to the target}$     \\
            GAN    & 15 degrees       & Neural network decides \\ \bottomrule
            \end{tabular}
        \end{table}
        
        \revised{To implement the aimbots we use the ViZDoom library to read the necessary information, such as the location of the enemy players on the screen. In a real-world scenario, this information would be obtained by reading the game memory or by using an object detection network (see Section \ref{sec:hacks-in-the-wild}). When the aimbot is activated, the code selects the closest enemy to the centre of the screen and provides the centre location of the enemy sprite (the white boxes in Figure \ref{fig:vizdoom_env}) to the aimbot as a target. The aimbot then moves the aim-point closer to this target. The aimbot does not fire the weapon.}
        
        The heuristic aimbot has a limited field of view and uses \say{slow aim}. The aimbot activates once a target is close enough to the player and to the crosshair (green and red boxes in Figure \ref{fig:vizdoom_env}). With the slow aim, the aimbot will not aim directly at the target in one frame; rather, it will move the mouse a portion of the distance to the target in each frame. Each movement by the aimbot includes a small amount of Gaussian noise to mimic the inaccuracy of human aiming, drawn from $\mathcal N(0, 0.2 \cdot d)$, where $d$ is the amount mouse is moved on the corresponding axis (in degrees). The multiplier $0.2$ was chosen with manual testing, where the value was increased until the aimbot's performance deteriorated.

        We use two aimbot variants in this setup: \textit{strong} and \textit{light} aimbots, where the stronger variant has a larger field of view and faster movement, and the lighter one more closely resembles existing hacks due to its smaller adjustments, which helps to avoid detection. The GAN-Aimbot uses the same field of view setting as the strong aimbot. These differences are detailed in Table \ref{tab:aimbot-settings}.

    \subsection{\revised{GAN-Aimbot setup}}
        \label{sec:training-gan}
        
        To create a discriminator in the GAN-Aimbot system, we use a neural network with two fully connected layers of $512$ units for the discriminator and two fully connected layers of $64$ units for the generator. The inputs and outputs of networks are described above in Section \ref{sec:gan-aimbot} and Figure \ref{fig:gan}. Both use exponential-linear unit (ELU) activation functions \cite{clevert2015fast}. We use a small network for the generator to speed up computation when the network is used as an aimbot. We found that ELU stabilised the training over rectified-linear units (ReLUs) and sigmoid function. We use hyper-parameters from \cite{arjovsky2017wasserstein}, setting the RMSprop learning rate to $5 \cdot 10^{-5}$, clipping weights to $0.01$ and updating the generator only every five training iterations. We use $c = 20$ steps for context size and $g = 5$ for the number of steps generated by the generator.
        
        \revised{We train the system} for $100$ epochs using a dataset of $20$ minutes of human gameplay, which was enough for losses to converge with a batch size of $64$. As the data used to train this system may change the aimbot's behaviour, we include two different systems in our experiments (Group 1 and Group 2), which use the same setup as above but with data from different players (See Section \ref{sec:data-collection12}). \revised{The goal is not to mimic an individual player but to mimic the dynamics of how humans control the mouse.}
        
\section{\revisedn{Comparing automatic classification systems (anti-cheats)}}
    \label{sec:classifier-comparison}
    
    \revisedn{We begin the experiments with two data collections and a comparison of different classifiers for automatic cheat detection. The goal is to find the strongest system, which we will then use for assessing aimbots' detection rates later.}
    
    \subsection{Deep neural network classifier system}
        \label{sec:classifier}
        To automatically detect hacking players, we follow the VACNet setup \cite{vacnet}, which uses a deep neural network (DNN) to detect players from mouse movement alone. A single feature vector is a tuple that contains mouse movement on both axes $0.5$s before and $0.25$s after a player has fired a weapon as well information on whether the shot hit the enemy. The final feature vector per shot is a tuple $(\Delta x_1, \ldots, \Delta x_n, \Delta y_1, \ldots, \Delta y_n, \text{is hit})$ with $n=25$. We only include feature vectors when the player is holding an instant-hit weapon; otherwise, the  \say{is hit} label does not coincide with mouse movement. All features apart from the \say{is hit} truth value are normalised by subtracting the mean and dividing by the standard deviation per dimension, computed over training data. Compared to other proposed machine-learning-based anti-cheat systems \cite{bayes_aimbot_detection, galli2011cheating, alayed2013behavioral}\removed{, which include information like enemy locations and keyboard buttons pressed}, these features focus on detecting aimbots but do not require game-specific modifications, and have been shown to be successful in practical applications \cite{vacnet}.
        
        The classifier network consists of two fully connected layers of size $512$, each followed by a ReLU activation. The final layer outputs logits for both classes. We use an L2-regularization weight of $0.01$ and train for $50$ epochs with batches of $64$ samples. These hyperparameters were obtained by manual trial-and-error to reach the same level of loss in training data and validation data ($10\%$ of samples from the training data). Network parameters are updated to minimise cross-entropy between true labels and predictions with gradient descent using an Adam optimiser \cite{adam} with a learning rate of $0.001$. To avoid biased training, we weight losses by an inverse probability $1 - p$ of the class being represented in a training batch. We also experimented with using ELU units as in the GAN-Aimbot, but this reduced the performance.
    
    \subsection{Baseline classifier systems}
        \removed{Neural networks are not the only classifiers, and} Instead of neural networks, previous work has explored using different methods, including support vector machines (SVM) \cite{alayed2013behavioral} and Bayesian networks \cite{bayes_aimbot_detection}. Motivated by this, we compare the DNN method to other binary classifiers using the auto-sklearn toolkit \cite{auto-sklearn} with the scikit-learn library \cite{scikit-learn}. This toolkit automatically tunes the hyperparameters of algorithms based on a withheld dataset, randomly sampled from the training data. 
        % Removed to save space
        We run experiments with a naive Bayes classifier, decision trees, random forests, linear classifiers, support vector machines (SVM) and linear discriminant analysis (LDA). 
        Each algorithm uses one-third of the training data for validation as recommended by the library documentation, and is then tuned for five hours on four 4.4Ghz CPU cores, which we found to be sufficient for generating stable results given the relatively small size of our datasets.
    
    \subsection{Data collection}
        \label{sec:data-collection12}
        The first data collection (Heuristic Collection) setup asked participants to play four five-minute games, where in the first two games no aimbot was enabled, the third game had a \textit{light} aimbot enabled and the fourth game had a \textit{strong} aimbot enabled. Two \textit{bona fide} games were used to balance the amount of \textit{bona fide} and hacking data during classification. \revisedn{We collected data from $23$ players; $18$ for the training set of the classifier and $5$ for the testing set. In total there were $8748$ negative (\textit{bona fide}) data points and $10732$ positive (cheating) points for training, and $2973$ negative and $3977$ positive points for testing.}
        
        We then took four players from the testing set and split them into two groups to train the Group 1 and 2 GAN-Aimbots. \revisedn{We used four players to train both GANs on a similar amount of data}. After training, we repeated the above data collection process (GAN Collection), where instead of using \textit{light} and \textit{strong} aimbots, players played with Group 1 and Group 2 GAN-Aimbots. We recruited the participants using the same channels but this time only included data from new participants to avoid mixing data of the same player in both training and testing sets. \revisedn{This resulted in fewer participants than in the first data collection by design, as otherwise there would have not been enough data to train the two GAN-Aimbots after train/test split. We collected data from $8$ players; $4$ for the training set and $4$ for the testing set. The training set had $2007$ negative and $2266$ positive points, and the testing set had $2074$ negative and $2126$ positive points. Note that this data is not used in this experiment, but it was recorded immediately after the first collection and will be used in Section~\ref{sec:detection-rates}.}

    \subsection{Results for classifier comparison}
        \begin{table}[t]
        \centering
        \caption{Balanced accuracies (\%) of different methods of detecting hackers.}
        \label{tab:classification-comparison}
        \begin{tabular}{lll}
            \toprule
            \textbf{Algorithm}             & \textbf{Train} & \textbf{Test}  \\ \midrule
            Naive Bayes       & 69.34 & 69.06 \\
            Decision Tree     & 80.47 & 72.09 \\
            LDA               & 62.91 & 60.23 \\
            SVM               & 89.40 & 84.34 \\
            Random Forest     & \textbf{94.54} & 77.33 \\
            Linear Classifier & 62.47 & 60.73 \\
            Neural Network    & 87.41 & \textbf{86.41} \\
            \bottomrule
        \end{tabular}
        \end{table}
        
        We compare classifiers with balanced accuracy, which is defined as the mean of individual accuracies $\frac{1}{2}(\text{acc}_\text{pos} + \text{acc}_\text{neg})$, with $\text{acc}_\text{pos}$ being the accuracy of positive samples and $\text{acc}_\text{neg}$ accuracy for negative samples.
        
        Table \ref{tab:classification-comparison} presents the balanced accuracies in the train and test sets of the Heuristic Collection. SVM, random forest and neural network shared similarly high performance in the testing set, which concurs with previous results \cite{alayed2013behavioral, vacnet}. We use the term \say{high} to describe the performance as this accuracy of $86\%$ is based on a small segment of mouse movement. Based on these results, we used the DNN system for the following experiments. \removed{and if we were to put this system to practical use, we could include data from entire games for more reliable results.} 

\section{Evaluating aimbot performance}
    \label{sec:aimbot-performance}

    \revisedn{Before assessing detection rates of aimbots, we study how much the GAN-Aimbot improves players' performance.}
    
    \subsection{Aimbot performance metrics}
        To measure the advantage aimbots give to the players, we track the accuracy and number of kills. Accuracy is the ratio of shots that hit an enemy. To measure accuracy, we only include shots fired with instant-hit weapons, as aimbots directly target the enemies and thus only help with instant-hit weapons.
    
    \subsection{Data collection for evaluating aimbot performance}
        In a real-world scenario, a cheater has to learn to play with their hack, as it may interfere with their behaviour. In our case, the aimbot takes control of the mouse abruptly when a target is close. For this reason, we created a more elaborate setup for aimbot performance evaluation (Performance Collection).
        
        Players first played $10$ minutes without an aimbot, then for $10$ minutes with a \textit{light} aimbot and finally played four games of $2.5$ minutes with the four aimbots (none, \textit{light}, \textit{strong} and GAN Group 1), in random order. The first two games are to ensure the player has sufficient time to learn game mechanics with and without an aimbot. We randomised the order of these last four games to reduce any bias from further learning during the final $10$ minutes. We only report the results of these final four games. \revisedn{Both of the trained GAN-Aimbots behaved in a similar way, hence we only included one of them}.
    
    \subsection{Results for aimbot performance evaluation}
        \begin{table}
            \centering
            \caption{Average ($\pm$ standard deviation) player accuracy and kills. Asterisks indicate a significant difference to ``None" according to two-tailed Welch's t-test ($N=22$, *$p < 0.05$, **$p < 0.01$, ***$p < 0.001$).}
            \label{tab:aimbot-performance}
            \begin{tabular}{lll}
                \toprule
                \textbf{Aimbot} & \textbf{Accuracy} & \textbf{Kills} \\ \midrule
                None            & $31.1 \pm 13.8$      & $12.1 \pm 5.1$\\
                Light           & $42.3 \pm 8.2$**     & $15.7 \pm 6.0$*\\
                Strong          & $56.1 \pm 8.7$***    & $22.0 \pm 7.5$***\\
                GAN             & $42.7 \pm 10.3$**    & $15.5 \pm 5.6$* \\
                \bottomrule
            \end{tabular}
        \end{table}

        Table \ref{tab:aimbot-performance} presents the average player performance metrics with and without different aimbots. All aimbots increased both metrics by a statistically significant amount. Since aimbots force players to aim at the enemy, slow-projectile weapons like rocket launchers and plasma guns become impractical to use (the player can only aim directly at the enemies, not in front of them), which explains the smaller increase in the number of kills, as these weapons are amongst the strongest. However, increased accuracy with the GAN-Aimbot indicates that our hack indeed works as an aimbot, making it a viable hack to use for cheating.

\section{Detection rates of the aimbots}
    \label{sec:detection-rates}
    
    \revisedn{In this section, we perform a detailed study of how easily the different aimbots can be detected. We use the data collected in the Heuristic Collection and the GAN Collection (Section~\ref{sec:data-collection12}).}

    \subsection{Detection rate metrics}
        To evaluate the DNNs performance in distinguishing different aimbots from humans, we use detection error tradeoff (DET) \cite{martin1997det} curves, which indicate the trade-off between false-positive (FPR) and false-negative rates (FNR) when the threshold is changed; a positive decision indicates that a player is flagged as a hacker. 
        %We are especially interested in maintaining low FPR, as game developers do not want to annoy \textit{bona fide} players by falsely accusing them of cheating.
        
        We also include equal error rates (EERs) and a detection cost function (DCF) \cite{doddington2000nist}. The EER is the error rate at the threshold where FPR equals FNR, which summarizes the DET curve in a single scalar value. The DCF incorporates two new parameters: the costs $C_\text{FP}$ and $C_\text{FN}$ for both error cases as well as a prior $p_\text{hacker} \in (0, 1)$, which reflects the assumed probability of encountering a hacking player. The DCF is then defined as \cite{doddington2000nist}
        \begin{equation}
            \label{eq:dcf}
            \text{DCF} = p_\text{hacker} C_\text{FN} P_\text{FN} + (1 - p_\text{hacker}) C_\text{FP} P_\text{FP},
        \end{equation}
        where $P_\text{FN}$ and $P_\text{FP}$ are FNR and FPR at a fixed threshold. The meaning of a positive/negative outcome is opposite to that of previous DCF literature, where a positive outcome is often a \say{good} thing (e.g., indicates a correct match, not a malicious actor). Our interpretation of the outcome is consistent with the previous cheat-detection literature, where a positive match indicates a cheating player \cite{galli2011cheating, bayes_aimbot_detection}.  
        
        For a single-scalar comparison, we report $\text{DCF}_\text{min}$, which corresponds to the minimal DCF obtained over all possible thresholds. This version of $\text{DCF}_\text{min}$ does not have an interpretable meaning, other than \say{lower is better}, but dividing this value by $\min \{p_\text{hacker} C_\text{FN}, (1-p_\text{hacker}) C_\text{FP}\}$ gives it an upper bound of one. If $\text{DCF}_\text{min} = 1.0$, the classifier is no better than always labelling samples as a hacker or \textit{bona fide}, depending on which has a smaller cost. All reported DCF values are normalised, minimum DCFs. 
        
        We set costs to unit costs $C_\text{FP} = C_\text{FN} = 1$ but vary the prior of hacking player and report the results at different levels $p_\text{hacker} \in \{0.5, 0.25, 0.1, 0.01\}$. We selected these values to cover a purely hypothetical situation in which half of the players are hackers but also more realistic scenarios where $1\%$ of players are hackers.\footnote{To provide a rough scale, the UnknownCheats forum reported 3.4M registered users in 2020 (user status is required to download hacks), while the Steam platform reported an average of 120M active monthly users in 2020~\cite{steam2020}.} We use unit-costs as we argue both error situations are equally bad: \revised{an undetected hacker can ruin the game of several players for a moment, but banning an innocent player can ruin the game for that individual until their appeal to lift the ban has been processed.}
        
        If these systems were applied to the real world, their performance would be lower than indicated by the  $\text{DCF}_\text{min}$. This is because this metric provides an optimal threshold, but in practice, finding the correct threshold may be difficult and the threshold likely changes over time \cite{brummer2010measuring}. 
    
    \subsection{Detection rate evaluation scenarios}
        \label{sec:scenarios}
        In a real-world scenario, game developers may or may not be aware of the hacks players are using. To reflect this, we design four different evaluation scenarios. Recall that there are two separate GAN-Aimbot systems trained using data from different players (Groups 1 and 2, Section \ref{sec:data-collection12}). The training and test sets are split player-wise, as further discussed in Section \ref{sec:data-collection12}.

        \textit{Worst-case scenario}: The hack is not known to the defender to any degree. In our case, this means that the GAN-Aimbot is a new type of aimbot and the anti-cheat is unaware of it. We train the classifiers only on data from heuristic aimbots and then evaluate using data from the two GAN-Aimbots. There is no worst-case scenario for \textit{light} and \textit{strong} aimbots, as these aimbots are commonly known at the time of writing.
        
        \textit{Known-attack}: The defender is familiar with the mechanisms of the aimbots and is trained to detect them. However, the defender does not know the precise parameters of the aimbot \removed{(the settings or weights of the network)}---only its general technique. This situation corresponds to the hack being public knowledge, but each hacker sets up their hack with different settings. In this scenario, we train a single classifier per aimbot type (four classifiers) and then evaluate the classifier using a similar but novel aimbot: the classifier trained using the Group 1 GAN is evaluated using Group 2 GAN data and vice versa, and the classifier trained using \textit{strong} aimbot data is evaluated using \textit{light} aimbot data and vice versa.
        
        \textit{Oracle}: The defender knows the exact GAN-Aimbot used by the hackers, including the network parameters. This corresponds to a hack being shared publicly on a hacking community; the defender can gather data on the hack and include it in the training set. To reflect this scenario, we train four classifiers as above but evaluate the classifiers using a test set of the same aimbot data.
        
        \textit{Train-on-test}: A scenario designed to test if the aimbot data is separable from \textit{bona fide} human data by including testing data in training. This scenario maximises accuracy but does not reflect a practical scenario. We train two classifiers, one for heuristic aimbots and one for GAN-Aimbots, and evaluate them using corresponding aimbots to ensure that there is enough data for training. The testing set is the same as in the previous scenarios. Note that this is a separate scenario from overfitting: \revised{we aim to determine how separable \textit{bona fide} data is from the aimbots' data.} \removed{we aim to determine an optimal decision boundary, but we do not want the network to memorise individual samples.} We observe that the training and validation losses stay roughly equal during training, which indicates that the network did not overfit.
    
    \subsection{Results for detection rates of aimbots}
        \label{sec:classification-results}
\begin{figure}[t]
            \begin{center}
            \centerline{\includegraphics[width=1.0\columnwidth]{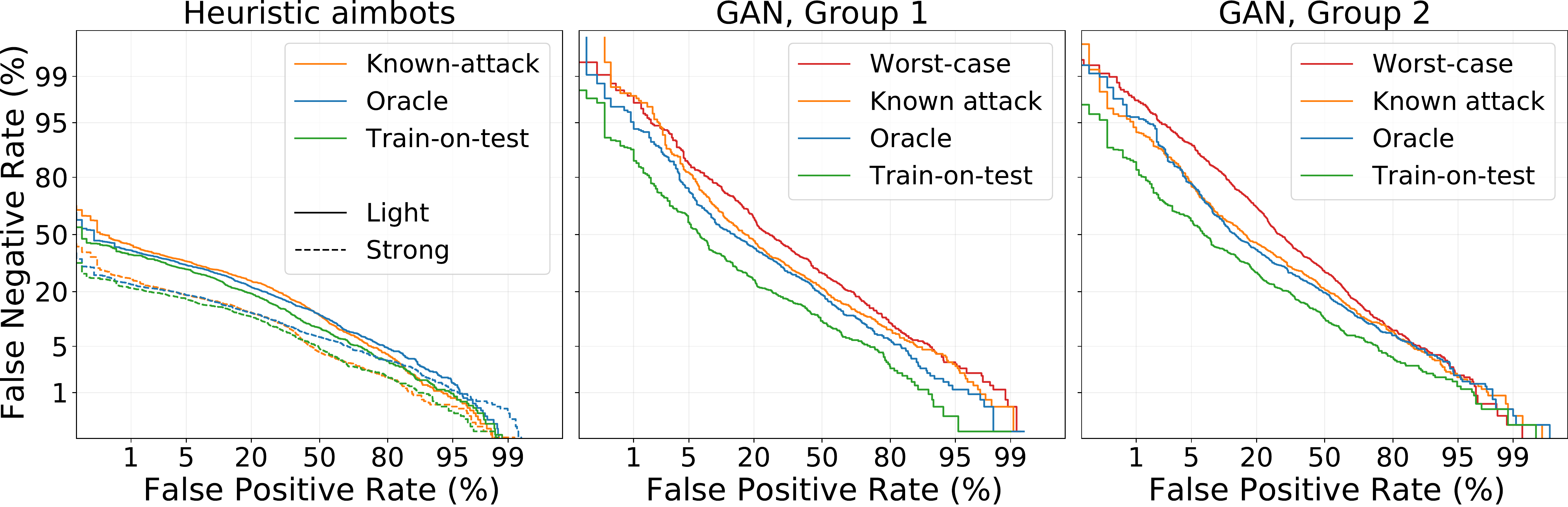}}
            \caption{DET curves of detecting different aimbots under different scenarios (Section \ref{sec:scenarios}) using DNN classifier (Section \ref{sec:classifier}). Curves towards the lower-left corner are better for security (easier to detect), while the worst possible case is a diagonal, descending line in the middle.}
            \label{fig:dets}
            \end{center}
        \end{figure}

        \removed{With the DNN classifier, we studied how easily different aimbots could be distinguished from \textit{bona fide} human play.} The DET curves in Figure \ref{fig:dets} highlight the major differences between the heuristic and the GAN aimbot in detectability: even when trained on the testing data, the FPR began to increase instantly when FNR decreased (left-most side in the figures). This result indicates that a large portion of samples from heuristic aimbots are separable from human data, while GAN-Aimbot samples are confused with \textit{bona fide} data. In more realistic scenarios, the error rates increased notably (note the normal-deviate scale of axis).
        
        Table \ref{tab:evaluation-results} presents the EER and $\text{DCF}_\text{min}$ values of these same scenarios. Although a GAN-Aimbot improved players' accuracy by an amount similar to a \textit{light} aimbot, the equal-error rates are notably higher than with the \textit{light} aimbot (i.e., GAN-Aimbot is harder to detect). When we consider the ratio of hackers to \textit{bona fide} players with $\text{DCF}_\text{min}$, we observe that the neural network classifier can be worse than not using it at all, even when testing data were contained in the training data. \revised{This is because the number of \textit{bona fide} players being flagged as hackers negated the benefit of occasionally detecting a hacker.}
        
        \begin{table}[t]
        \centering
        \caption{Evaluation results of detecting aimbots, with lower values meaning better for the anti-cheat. Bolded $\text{DCF}_\text{min}$ values are above $0.95$, meaning the classifier caused almost as much harm as not using it. EERs are in percentages (\%).}
        \label{tab:evaluation-results}
        \setlength\tabcolsep{5pt}
\begin{tabular}{llccccc}
\toprule
\multirow{2}{*}{\textbf{Aimbot}}                      & \multirow{2}{*}{\textbf{Scenario}}                & \multirow{2}{*}{\textbf{EER}}               & \multicolumn{4}{c}{$\text{DCF}_\text{min}, \; p_\text{hacker} = \ldots$}                                                            \\
\multicolumn{1}{l}{}          & \multicolumn{1}{l}{} & \multicolumn{1}{l}{} & $0.5$ & $0.25$ & $0.1$ & $0.01$ \\ \midrule
\multirow{4}{*}{Light}          & Worst-case     & -            & -            & -            & -            & - \\
                                & Known-attack   & 21.06          & .3713         & .4099         & .4494         & .5972 \\
                                & Oracle         & 21.91          & .3846         & .4344         & .4857         & .6761 \\
                                & Train-on-test  & 19.02          & .3365         & .3957         & .4549         & .5386 \\ &&& \\
\multirow{4}{*}{Strong}         & Worst-case     & -            & -            & -            & -            & - \\
                                & Known-attack   & 14.65          & .2300         & .2543         & .2974         & .4635 \\
                                & Oracle         & 13.27          & .2226         & .2507         & .2801         & .3985 \\
                                & Train-on-test  & 12.55          & .2052         & .2376         & .2793         & .3379 \\ &&& \\
\multirow{4}{*}{GAN1}       & Worst-case     & 41.92          & .8380         & .9990         & \textbf{1.0000}         & \textbf{1.0000} \\
                                & Known-attack   & 30.03          & .5779         & .8016         & .9236         & \textbf{1.0000} \\
                                & Oracle         & 28.27          & .5388         & .7499         & .9393         & \textbf{.9869} \\
                                & Train-on-test  & 19.40          & .3795         & .6114         & .8379         & \textbf{.9785} \\ &&& \\
\multirow{4}{*}{GAN2}       & Worst-case     & 38.31          & .7651         & .9978         & \textbf{.9989}& \textbf{.9991} \\
                                & Known-attack   & 33.21          & .6200         & .7901         & .9168         & \textbf{.9735} \\
                                & Oracle         & 25.90          & .4955         & .6839         & .8199         & \textbf{.9672} \\
                                & Train-on-test  & 19.55          & .3805         & .5788         & .7578         & .9276 \\ \bottomrule
\end{tabular}
\end{table}
    
    \subsection{\revisedn{Impact of player skill on detection rate}}
        \revisedn{To study if the player's skill in the game affects the detection rate of the aimbot, we computed the correlation between players' \textit{bona fide} performance (kills and accuracy) and the average detection scores, using data from the Performance Collection in the oracle scenario. We found virtually no correlation when GAN-Aimbot was used, neither with kills nor accuracy ($-0.087$ and $-0.037$, respectively). Note that, because of how the collection was performed, data of the same player may have been used to train the classifier and in this evaluation, thus skewing the results. As such, these results are preliminary.}
    
    \subsection{Analysis of mouse movement}
        \begin{figure}[t]
            \begin{center}
            \centerline{\includegraphics[width=1.0\columnwidth]{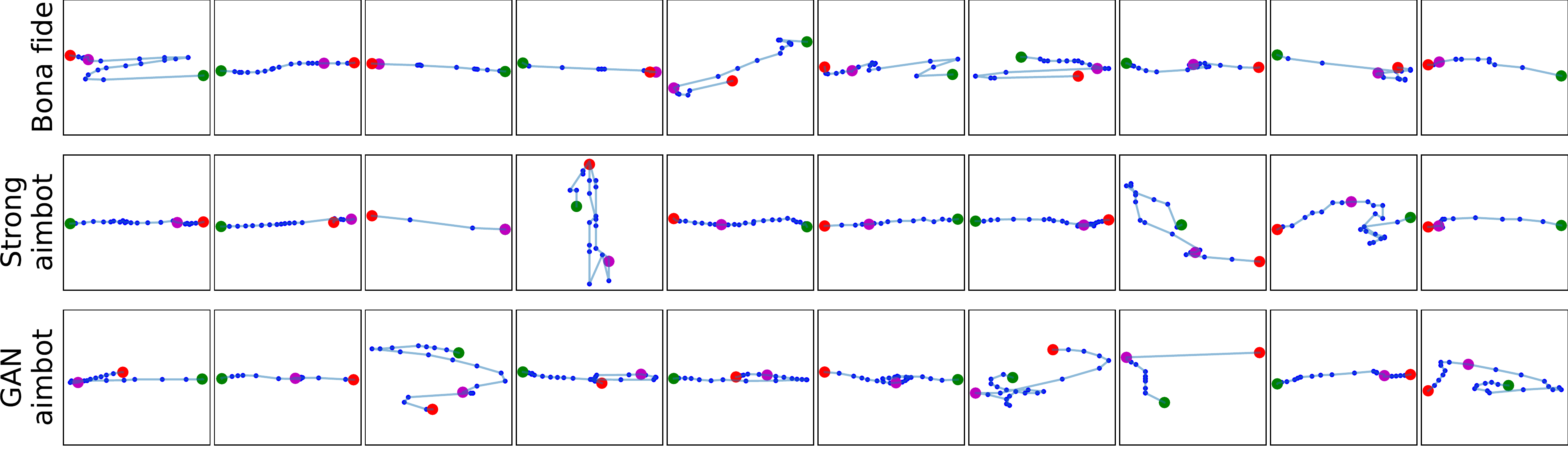}}
            \caption{Randomly selected examples of mouse trajectories. Green marks the beginning of the trajectory, purple is the point where the player fires the gun and red is where the trajectory ends. Axis ranges are not shared between figures but are equal per figure (aspect ratio is one).}
            \label{fig:trajectories}
            \end{center}
        \end{figure}
        
        \begin{table}[!t]
        \centering
        \caption{\revisedn{Statistics of the mouse movement with different aimbots: average mouse movement per axis, Pearson correlation between movement per axis and correlation between successive steps per axis. The first three rows are computed on absolute units.}}
        \label{tab:mouse-movement}
        \begin{tabular}{lccc}
            \toprule
            \textbf{Measure} & \textbf{None} & \textbf{Heuristic} & \textbf{GAN} \\ \midrule
            Avg. yaw           & $1.47 \pm 3.07$    & $1.70 \pm 3.12$ & $1.58 \pm 3.08$  \\
            Avg. pitch         & $0.13 \pm 0.27$    & $0.25 \pm 0.58$ &  $0.19 \pm 0.29$   \\
            Axis corr.         & $0.392$            & $0.255$ &  $0.342$   \\
            Step corr. (yaw)   & $0.742$            & $0.749$ &  $0.728$   \\
            Step corr. (pitch) & $0.658$            & $0.042$ &  $0.648$   \\
            \bottomrule 
        \end{tabular}
        \end{table}
        
        %Raw evaluation metrics seem promising for the GAN-Aimbot, however the numbers do not tell use \textit{how} the learned aimbot functions and how it is able to fool the classifier. 
        Since the classifier was trained to distinguish \textit{bona fide} players from hacking players, one could assume that an aimbot that evades detection behaves like a \textit{bona fide} player. This is not necessarily true, however, as the machine learning aimbot could have simply found a weakness in the classification setup akin to adversarial attacks \cite{goodfellow2014explaining}.
        
        Figure \ref{fig:trajectories} outlines how the GAN-Aimbot moves the mouse during gameplay; the network has learned to move in small steps. If one views videos of the gameplay, GAN-Aimbot also occasionally \say{overshoots} the target point and \revisedn{intentionally misses the target, unlike heuristic aimbots that stick to the target}.
        
        Looking at the average, absolute per-step mouse movement in Table \ref{tab:mouse-movement} (first and second rows), we see that the heuristic aimbots had the most movement on the vertical axis (pitch), followed by GAN-Aimbots and finally by \textit{bona fide} players. This result is expected, as moving the PC mouse vertically is less convenient than moving it horizontally (give it a try). Heuristic aimbots may not take this into account, but the GAN-Aimbot learned this restriction from human data. The same can be seen in the correlation of the two variables (third row, significantly different from zero with $p < 0.001$), where no aimbot and GAN aimbots move the mouse in both axes at the same time more than the heuristic aimbot does.
        
        \revisedn{Finally, the last two rows show the correlation between the successive mouse movements, measured as $\text{Corr}(\Delta y_{t+1}, \Delta y_t)$, averaged over all data points. Heuristic aimbots also have a near-zero correlation on the vertical axis. If the target moves closer or further away from the player, the target point moves vertically on the screen, and heuristic aimbots adjust accordingly. GAN-Aimbot has learned to avoid this.}
        
        \removed{A similar pattern can be seen in the correlations between absolute per-step movement on the two axes, where \textit{bona fide} data had a Pearson correlation of $0.39$; heuristic aimbots $0.25$; and GAN-Aimbots, $0.34$ (computed over the first two data collections, all significantly different from zero with a two-tailed $p < 0.001$). This result signals that \textit{bona fide} players move more often along both axes simultaneously, whereas heuristic aimbots move more along a single axis at any time.}
    
    \subsection{Classifying whole games with multiple feature vectors}
        \label{sec:game-classification}
        \begin{figure}[t]
            \begin{center}
            \centerline{\includegraphics[width=0.7\columnwidth]{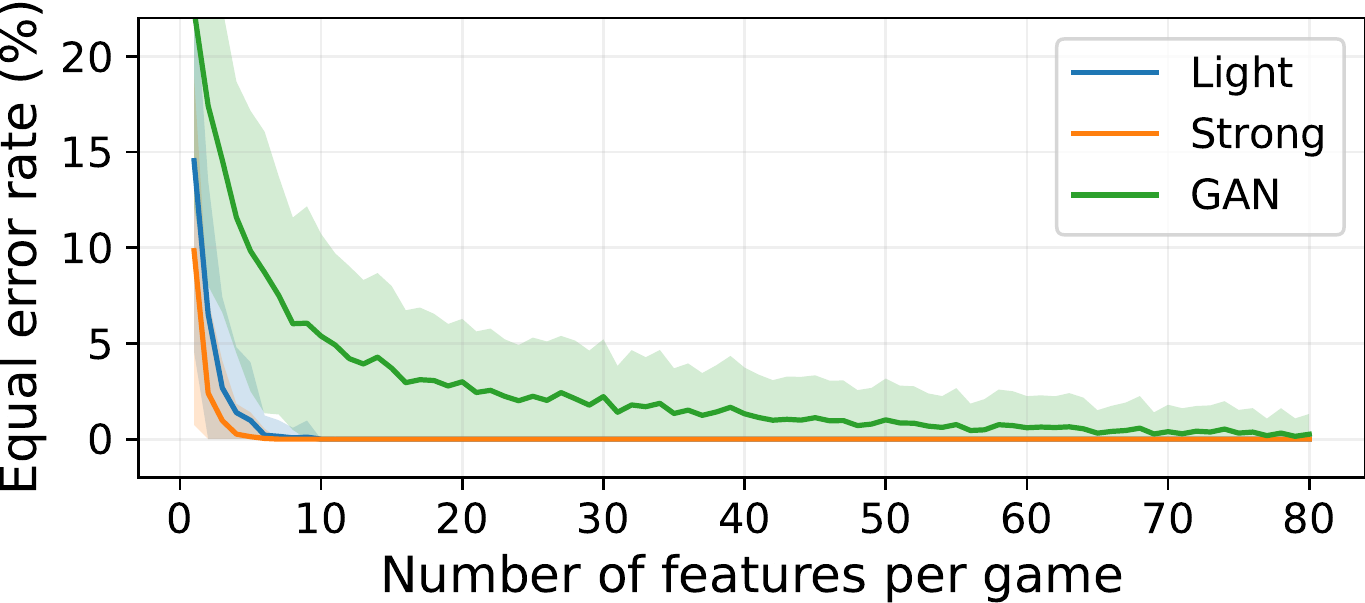}}
            \caption{\revisedn{EERs of classifying whole games, using a varying number of randomly sampled feature vectors (x-axis) and averaging the score over all these vectors. The line is the average EER over $200$ repetitions, and the shaded region is plus/minus one standard deviation.}}
            \label{fig:multi-vector-classification}
            \end{center}
        \end{figure}
        
        \revisedn{While classifying individual feature vectors simplifies the experiments, practical applications may aggregate data over multiple events. To study how this affects classification results, we use the classifiers trained in the Oracle scenario (Section~\ref{sec:scenarios}) to score individual feature vectors and then average the scores per game. We then compute the EER (\textit{bona fide} vs. cheating)  of classifying whole games based on these averaged scores. We repeat this for a different number of feature vectors, randomly sampling the vectors per game, and repeat the process $200$ times to stabilize the noise from sampling.}
        
        \revisedn{Figure~\ref{fig:multi-vector-classification} show the results. \textit{Light} and \textit{strong} aimbots are easily detected with less than ten feature vectors (roughly 7 seconds of game time). GAN-Aimbot becomes increasingly detectable with more data, approaching zero EER but not reaching it reliably. This indicates that the correct aggregation of detection scores over individual events will likely work as a good detection mechanism, assuming the GAN-Aimbot cheat is available to the anti-cheat developers.}

\section{Evaluating human detection rates}
    \revisedn{In addition to automatic detection systems, we are interested in whether humans are able to distinguish users of aimbot from \textit{bona fide} players by watching video replays of the players or not.}
    
    \subsection{Data collection for video recordings}
        To gather video recordings of gameplay \revisedn{(Video Collection)}, we organised a data collection procedure similar to the Heuristic Collection and GAN Collection with different aimbots (none, \textit{light}, \textit{strong} and GAN Group 1), in a randomised order to avoid constant bias from learning the game mechanics. We asked five participants from the previous data collection process (GAN Collection) to participate as they were already familiar with the setup, and had these participants record videos of their gameplay.
        
        We then obtained three non-overlapping clips $30$s in length at random from each recording, where each clip contained at least one kill. In total, we collected $60$ video clips, $15$ per aimbot including the no aimbot condition.
    
    \subsection{Setup for collecting human grading}
        We let human judges grade the recordings \revisedn{(Grade Collection)} with one of the following grades and instructions:

        \begin{enumerate}
            \item Not suspicious. I would not call this player a cheater.
            \item Suspicious. I would ask for another opinion and/or monitor this player for a longer period of time to determine if they were truly cheating.
            \item Definitely cheating. I would flag this player for cheating and use the given video clip as evidence.
        \end{enumerate}
        
        The above grading system was inspired by situations in the real world. In the most obvious cases, a human judge can flag the player for cheating (Grade 3). However, if the player seems suspicious (e.g., fast and accurate mouse movements but not otherwise suspicious), they can ask for a second opinion or study other clips from the player (Grade 2). \removed{Sometimes the clips a human judge receives may come from genuinely \textit{bona fide} players, which can happen when an automatic system falsely marks a player as suspicious and asks a human to verify it.}
        
        \revised{We include both experienced judges who have experience in studying gameplay footage in a similar setup and FPS players with more than a hundred hours of FPS gameplay performance.} We ensure that all participants are familiar with the concept of aimbots and how to detect them. \revised{We provide judges with a directory of files and a text file to fill and submit back, without time constraints.}
        \revisedn{We obtained the grades from three experienced judges and three FPS players}.
    
    \subsection{Results of human evaluation}
        \label{sec:human-evaluation-results}
        \begin{figure}[t]
            \centering
            \includegraphics[width=0.6\columnwidth]{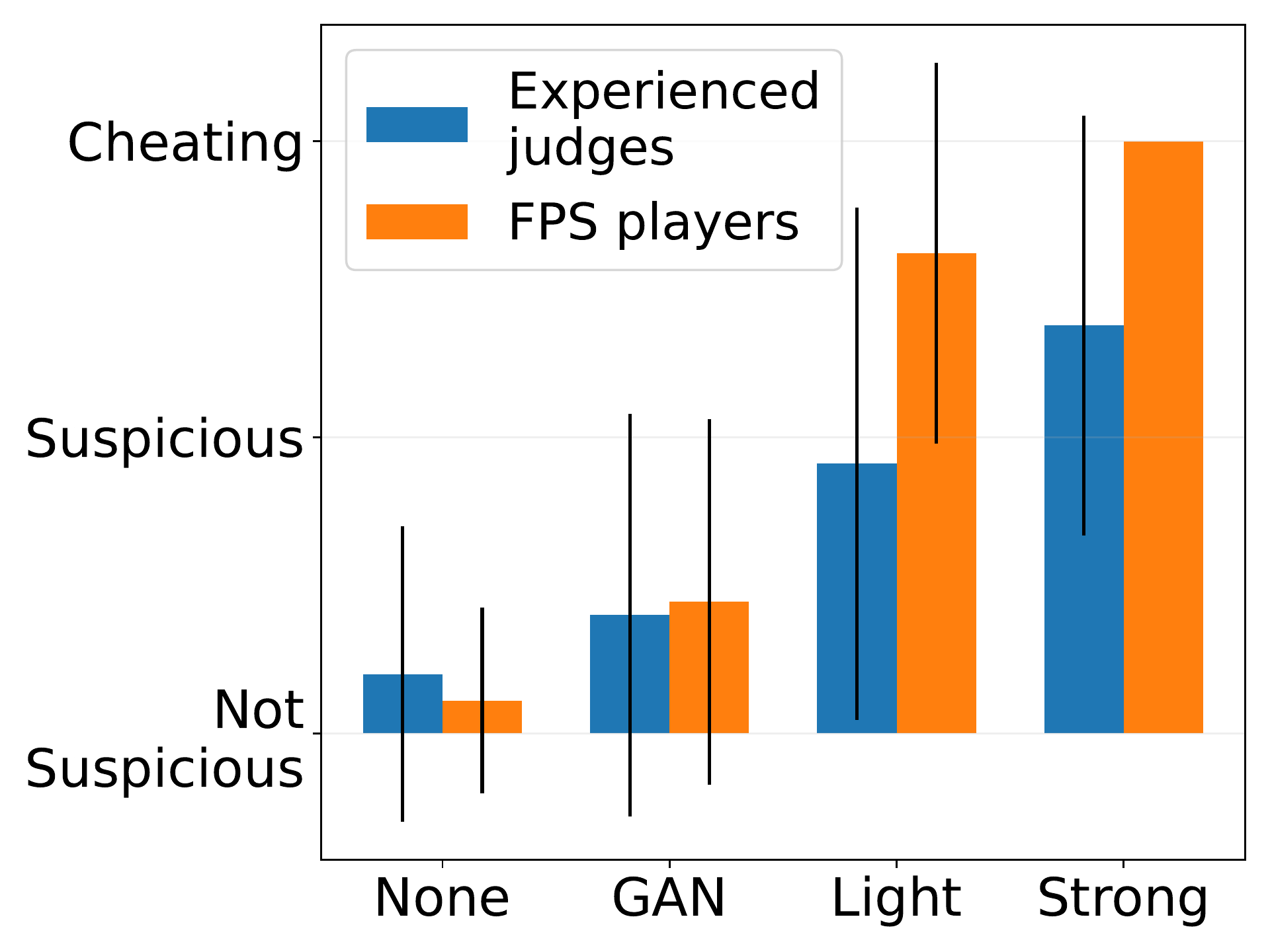}
            \caption{Average grading (y-axis) given by human graders for different aimbots (x-axis), with a black line representing plus-minus one standard deviation, computed over $90$ samples. \revisedn{For \textit{strong} aimbot, all FPS player judges voted cheating, hence the standard deviation is zero.}}
            \label{fig:human-grading}
        \end{figure}
        
        \begin{table}[t]
        \centering
        \caption{Ratio of grades per aimbot per grader group, in percentages (\%).}
        \label{tab:human-grading-ratios}
        \begin{tabular}{lcccc}
            \toprule
            \textbf{Experienced judges} & \textbf{None} & \textbf{GAN}  & \textbf{Light} & \textbf{Strong} \\ \midrule
            Not suspicious     & \revisedn{84.4} & 71.1 & 42.2 & 13.3    \\
            Suspicious         & 11.1 & 17.8 & 24.4 & 35.6    \\
            Cheating           & 4.4  & 11.1 & 33.3 & 51.1    \\ \midrule
            \textbf{FPS players} &  &   &  &  \\ \midrule
            Not suspicious & 88.9 & 62.2 & 8.9   & 0.0    \\
            Suspicious     & 11.1 & 31.1 & 20.0  & 0.0    \\
            Cheating       & 0.0  & 6.7  & 71.1  & 100.0 \\
            \bottomrule
        \end{tabular}
        \end{table}
        
        The results of the human evaluation are presented in Figure \ref{fig:human-grading} and Table \ref{tab:human-grading-ratios}. Much like with automatic cheat detection, the GAN-Aimbot was less likely to be classified as suspicious or definitely cheating, while the \textit{strong} aimbot was always correctly detected by FPS players with some hesitation from the experienced judges. The latter group commented that cheating had to be exceptionally evident from the video clip to mark the player a cheater, as these judges were aware of \textit{bona fide} human players who may appear inhumanly skilled.
        
%     \begin{table}[t]
%     \centering
%     \caption{Statistics of all the collected data. Entries with a slash show the split between the training and testing set, respectively. For each aimbot, we report the number of feature vectors (Section \ref{sec:classifier}) \todo{where to move the training statistics?}}
%     \label{tab:data-collected}
%     \setlength\tabcolsep{5pt}
% \begin{tabular}{lccccc}
% \toprule
% \textbf{Measure}      & \textbf{1}     & \textbf{2}     & \textbf{3}     & \textbf{4}      & \textbf{5}        \\ \midrule
% Data type & Data  & Data  & Data  & Videos & Grades \\
% Participants   & 18 / 5    & 4 / 4     & 22     & 5      & 6        \\
% Game time & 7.7h  & 2.7h  & 10.5h  & 1.6h   & N/A      \\ \midrule
% \textbf{Aimbots} \\ \midrule
% None              & 8748 / 2973 & 2007 / 2074 & N/A  & N/A    & N/A      \\ 
% \textit{Light}    & 5042 / 1864 & N/A & N/A  & N/A    & N/A      \\ 
% \textit{Strong}   & 5690 / 2113 & N/A & N/A  & N/A    & N/A      \\ 
% GAN (G1)       & N/A & 1127 / 1069 & N/A  & N/A    & N/A      \\ 
% GAN (G2)       & N/A & 1139 / 1057 & N/A  & N/A    & N/A      \\ \bottomrule

% \end{tabular}
%     \end{table}

\section{Discussion}
    \label{sec:discussion}
    \removed{The results are discouraging for the future of multiplayer game developers: while statistical analysis and machine learning can improve anti-cheats \removed{considerably more than heuristic rules}, hackers can employ the same techniques, and our results demonstrate that hackers may hold the current advantage. After training with less than an hour of human gameplay data, the GAN-Aimbot acted as a functional aimbot yet was hard to distinguish from human players, both by an automatic system and by human judges.}
    \revised{The results provide direct answers for our RQs: the data-based aimbot does improve players' performance (RQ1) while remaining undetectable (RQ2). This indicates that the risk of data-based aimbots to the integrity of multiplayer video games exists, especially considering that the evaluated method only required less than an hour's worth of gameplay data to train.}
    
    Beyond the field of gaming, the same human-like mouse control could be used to attack automated Turing tests, or CAPTCHAs \cite{von2004telling}, which study a user's inputs to determine whether a user is a bot. Previous work has already demonstrated effective attacks against image-based CAPTCHAs \cite{zi2019end} and systems that analyse the mouse movement of the user \cite{akrout2019hacking}. The latter work simulated the CAPTCHA task and trained a computer agent to behave in a human-like way, whereas our system only required human data to be trained.
    
    However, our results have some positive outcomes. As the trained aimbot was harder to distinguish from human players, it is reasonable to assume its behaviour was closer to human players. Such a feature is desired in so-called aim-assist features of video games, where the game itself helps player's aim by gently guiding them towards potential targets \cite{vicencio2014effectiveness}. Our method could be used for such purposes. As it does not set restrictions on types of input information, it could also be a basis for a more generic game-assist tool that game developers could include in their games to make learning the game easier. Alternatively, this methodology could be used to create human-like opponents to play against, similar to the 2K BotPrize competition~\cite{philip2009botprize}, where participants were tasked to create computer players that were indistinguishable from human players in \textit{Unreal Tournament}. Such artificial opponents could be more fun to play against as they could exhibit human flaws and limitations. 
    
    \subsection{Possible future directions to detect {GAN-Aimbots}}
        Let us assume that hacks similar \revisedn{to GAN-Aimbot} become popular and hacking communities share and improve them openly. How could these hacks be prevented by game developers? \revisedn{Results of Section~\ref{sec:detection-rates} indicate that GAN-Aimbots evade detection, so we need new approaches to detect hackers. The following are broad suggestions that apply to GAN-Aimbots but could be also applied to detect other forms of cheating.}
        
        An expensive but ideal approach is to restrict hackers' access to the game's \revisedn{input and output} in the first place (e.g., game consoles that prevent the user from controlling the game character by means other than the gamepad). An enthusiastic hacker could still build a device to mimic that gamepad and send input signals from program code---or even construct a robot to manipulate the original gamepad---but we argue that such practices are not currently a major threat, as many players would not want to exert so much effort just to cheat in a video game.
        
        \revised{Another approach would be to model players' behaviour over time}. Sudden changes in player behaviour could be taken as a sign of potential hacking and the player could be subjected to more careful analysis. The FairFight anti-cheat advertises similar features~\cite{fairfight}.
        % Of course, this could very well mean somebody else is playing on their account, like sharing a game with your friend. A player could also create a new account and always play with the hack, in which case the behaviour of the hack is included in the player's normal behaviour and nothing suspicious is detected.
        
        Extending the idea above, anti-cheat players could create \say{fingerprints} of different hacks, similar to byte-code signatures already used to detect specific pieces of software running on a machine. These fingerprints would model the behaviour of the hack, not that of the players. \removed{Much like biometrics can be used to distinguish different people \cite{jain2006biometrics}, these fingerprints could be used to identify different hacks.} However, this process would require access to hacks to build such a fingerprint. 
        
        %\todo{highlight these}

        Finally, anti-cheats could extend beyond just modelling the gameplay behaviour of the player in terms of device inputs and instead include behaviour in social aspects of the game. Recent work has indicated that the social behaviour of hackers (e.g., chatting with other players) differs notably from that of \textit{bona fide} players~\cite{bernardi2017game}, and combined with other factors, this information could be used to develop an anti-cheat that could reliably detect hackers with and provide an explanation of why this person player was flagged \cite{explainable-cheating}.
    
    \subsection{Limitations of this work}
        We note that all experiments were run using one type of classifier and a single set of features, both of which could be further engineered for better classification accuracy. However, the GAN-Aimbot could also be further engineered, and the discriminator element could be updated to match the classifier. Theoretically, the adversarial-style training should scale up to any type of discriminator, until the generator learns to generate indistinguishable mouse movement from \textit{bona fide} players' data. We also used a fixed target inside an enemy player sprite, but a human player is unlikely to always aim at one very specific spot. This target point could also be determined via machine learning \revisedn{to appear more human-like}.
        
        The experiments were conducted using a single game environment and focused solely on mouse movement. While the results might apply to other FPS games, they may not necessarily generalise well to other games and other input mechanisms. Players of a real-time strategy game (e.g. Starcraft II) could theoretically \revised{design a system to automatically control the units}, for example. \removed{but this would require one to predict what buttons on the keyboard should be pressed and when, and other games involving more complicated control than just pointing and clicking would need more engineering to develop a functional algorithm.}

\section{\revised{Broader impact}}
    \label{sec:impact}
    We detail and present a method that, according to our results, achieves what was deemed negative (a cheat that works and is undetectable). A malicious actor could use the knowledge of this work to implement a similar hack in a commercial game and use it. For this reason, we have to assess the broader impact of publishing this work~\cite{cook2021social}, for which use the framework of Hagendorff (2021) \cite{hagendorff2021forbidden}. This decision framework, based on similar decision frameworks in biological and chemical research, is designed to address the question of whether the technology could be used in a harmful way by defining multiple characteristics of the harm. We answer each of these with a \say{low} or \say{high} grading, where \say{high} indicates a high chance of harm or a high amount of harm.
    
    \paragraph{Magnitude of harm.} Low. \revisedn{The focus is on the individuals. This work could lead to players disliking games as new hackers emerge, to erosion of trust in the video game communities and to monetary costs to companies. However, no directed harm is done to individuals (e.g., no bodily harm, no oppression), and the nuisance is limited to their time in video games. Cheating is already prevalent (Section~\ref{sec:hacks-in-the-wild}), and while this knowledge could make it more widespread, it would not be a sudden, rapid change in the video game scene.}
    
    \paragraph{Imminence of harm.} High. With enough resources and IT skills, a malicious actor can use the knowledge of our study for harm.
    
    \paragraph{Ease of access.} Low chance of harm. The method does not require specialized hardware but requires setting up a custom data collection. There are also no guarantees that the described method would work as is in other games.
    
    \paragraph{Skills required to use this knowledge} Low chance of harm. Using this knowledge for malicious purposes requires knowledge in machine learning and programming. The included source code only works in the research environment used here and can not be directly applied to other games. With all the likelihood, this will also require  training and potentially designing new neural network architectures with associated practical training recipes.
    
    \paragraph{Public awareness of the harm.} High (hidden) impact of the harm. Given the relative simplicity of this method and the number of active hacking communities, it is possible that a private group of hackers have already used a similar method to conceal their cheating. This would make this work more important, as now the defenders (anti-cheat developers, researchers) are also aware of the issue and can work towards more secure systems.
    
    \revisedn{While in the short term sharing this work may support cheating communities (unless it has already been used secretly), we believe it will be a net-positive gain as researchers and game developer companies improve their anti-cheat systems to combat this threat. If new threats are gone unnoticed, the defence systems may be ill-designed to combat them; our system can detect heuristic aimbots reliably with less than ten feature vectors (seven seconds of game time), but GAN-Aimbot requires more than $70$ (one minute), as illustrated by results in Section~\ref{sec:game-classification}. As security is an inherently adversarial game whereby novel attacks are being counteracted by new defences (and vice versa), we hope our study supports the development of new defences by providing information on emerging attacks.}
    
    \revisedn{In the short-term future, video game developers could employ stricter traditional methods (Section~\ref{sec:anti-cheats}) for sensitive games, such as requiring a phone number for identification. For data-based solutions, results in Section~\ref{sec:game-classification} indicate that aggregating results over multiple events can still work as a detection measure, though detection of GAN-Aimbots was found to require more data for reliable detection. For more sustainable solutions, however, options described in Section~\ref{sec:discussion} should be explored.}

    %While in the short term sharing this work may cause harm (a malicious user adapts this method), we believe it will be a net-positive gain as researchers and companies adapt their anti-cheat systems for this threat. This work increases the research community awareness of the issue, which can lead to further work into improving this situation. In addition, this serves to educate the public of the possible future where artificial players are indistinguishable from human players, which may lead to large changes in the video gaming community, in addition to other benefits discussed in Section \ref{sec:discussion}. By publicly sharing these results the community may already start discussing potential directions, should such future happen.
    \color{black}

\section{Conclusion}
    In this work, we covered the current state of cheating in multiplayer video games and common countermeasures used by game developers to prevent cheating. Some of these countermeasures use machine learning to study player behaviour to spot cheaters, but we argue that cheaters could do the same and use machine learning to devise cheats that behave like humans and thus avoid detection. To understand this threat, we designed a proof-of-concept method of such an approach and confirmed that the proposed method is both a functional cheat and also evades detection by both an automatic system and human judges. We share these results, as well as the experimental code and data (at \url{https://github.com/Miffyli/gan-aimbots}), to support and motivate further work in this field to keep video games trustworthy and fun to play for everyone.

%\section*{Acknowledgments}
%    We thank Rosa Gonz\'alez Hautam\"aki for comments on the data collection setups.


\bibliographystyle{IEEEtran}
\bibliography{references}

\begin{thebibliography}{10}
\providecommand{\url}[1]{#1}
\csname url@samestyle\endcsname
\providecommand{\newblock}{\relax}
\providecommand{\bibinfo}[2]{#2}
\providecommand{\BIBentrySTDinterwordspacing}{\spaceskip=0pt\relax}
\providecommand{\BIBentryALTinterwordstretchfactor}{4}
\providecommand{\BIBentryALTinterwordspacing}{\spaceskip=\fontdimen2\font plus
\BIBentryALTinterwordstretchfactor\fontdimen3\font minus
  \fontdimen4\font\relax}
\providecommand{\BIBforeignlanguage}[2]{{%
\expandafter\ifx\csname l@#1\endcsname\relax
\typeout{** WARNING: IEEEtran.bst: No hyphenation pattern has been}%
\typeout{** loaded for the language `#1'. Using the pattern for}%
\typeout{** the default language instead.}%
\else
\language=\csname l@#1\endcsname
\fi
#2}}
\providecommand{\BIBdecl}{\relax}
\BIBdecl

\bibitem{blizzard2020}
``Activision blizzard's second quarterly results 2020,''
  \url{https://investor.activision.com/static-files/926e411c-7d05-49f8-a855-6dc7646e84c4},
  accessed February 2022.

\bibitem{mpgh}
``Multiplayer game hacking,'' \url{https://mpgh.net}, accessed February 2022.

\bibitem{uc}
``Unknowncheats,'' \url{https://unknowncheats.me}, accessed February 2022.

\bibitem{yan2005systematic}
J.~Yan and B.~Randell, ``A systematic classification of cheating in online
  games,'' in \emph{Proceedings of 4th ACM SIGCOMM workshop on Network and
  system support for games}.\hskip 1em plus 0.5em minus 0.4em\relax ACM, 2005,
  pp. 1--9.

\bibitem{gamegenie}
M.~Nielsen, ``Game genie - the video game enhancer,''
  \url{http://www.nesworld.com/gamegenie.php}, accessed February 2022, 2000.

\bibitem{galli2011cheating}
L.~Galli, D.~Loiacono, L.~Cardamone, and P.~L. Lanzi, ``A cheating detection
  framework for unreal tournament iii: A machine learning approach,'' in
  \emph{Conference on Computational Intelligence and Games}.\hskip 1em plus
  0.5em minus 0.4em\relax IEEE, 2011, pp. 266--272.

\bibitem{alayed2013behavioral}
H.~Alayed, F.~Frangoudes, and C.~Neuman, ``Behavioral-based cheating detection
  in online first person shooters using machine learning techniques,'' in
  \emph{Conference on Computational Intelligence and Games}.\hskip 1em plus
  0.5em minus 0.4em\relax IEEE, 2013, pp. 1--8.

\bibitem{bayes_aimbot_detection}
S.~F. Yeung and J.~C. Lui, ``Dynamic bayesian approach for detecting cheats in
  multi-player online games,'' \emph{Multimedia Systems}, vol.~14, no.~4, pp.
  221--236, 2008.

\bibitem{karras2018style}
T.~Karras, S.~Laine, and T.~Aila, ``A style-based generator architecture for
  generative adversarial networks, 2019 ieee,'' in \emph{Computer Vision and
  Pattern Recognition}, 2018, pp. 4396--4405.

\bibitem{sizov2015joint}
A.~Sizov, E.~Khoury, T.~Kinnunen, Z.~Wu, and S.~Marcel, ``Joint speaker
  verification and antispoofing in the $ i $-vector space,'' \emph{IEEE
  Transactions on Information Forensics and Security}, vol.~10, no.~4, pp.
  821--832, 2015.

\bibitem{ho2016generative}
J.~Ho and S.~Ermon, ``Generative adversarial imitation learning,'' in
  \emph{Advances in neural information processing systems}, 2016, pp.
  4565--4573.

\bibitem{fu2018learning}
J.~Fu, K.~Luo, and S.~Levine, ``Learning robust rewards with adversarial
  inverse reinforcement learning,'' in \emph{International Conference on
  Learning Representations}, 2018.

\bibitem{goodfellow2014generative}
I.~Goodfellow, J.~Pouget-Abadie, M.~Mirza, B.~Xu, D.~Warde-Farley, S.~Ozair,
  A.~Courville, and Y.~Bengio, ``Generative adversarial nets,'' in
  \emph{Advances in neural information processing systems}, 2014, pp.
  2672--2680.

\bibitem{ratha2001enhancing}
N.~K. Ratha, J.~H. Connell, and R.~M. Bolle, ``Enhancing security and privacy
  in biometrics-based authentication systems,'' \emph{IBM systems Journal},
  vol.~40, no.~3, pp. 614--634, 2001.

\bibitem{roberts2007biometric}
C.~Roberts, ``Biometric attack vectors and defences,'' \emph{Computers \&
  Security}, vol.~26, no.~1, pp. 14--25, 2007.

\bibitem{redmon2016you}
J.~Redmon, S.~Divvala, R.~Girshick, and A.~Farhadi, ``You only look once:
  Unified, real-time object detection,'' in \emph{Computer Vision and Pattern
  Recognition}, 2016, pp. 779--788.

\bibitem{battleye}
B.~Innovations, ``Battleye,'' \url{https://battleye.com/}, accessed February
  2022.

\bibitem{punkbuster}
Even-Balance, ``Punkbuster,'' \url{https://evenbalance.com}, accessed February
  2022.

\bibitem{fairfight}
GameBlocks, ``Fair{F}ight,'' \url{https://www.i3d.net/products/hosting/
  anti-cheat-software}, accessed February 2022.

\bibitem{eac}
Kamu, ``Easy {A}nti-{C}heat,'' \url{https://easy.ac}, accessed February 2022.

\bibitem{vacnet}
J.~McDonald, ``Using deep learning to combat cheating in {CSGO},'' GDC 2018,
  \url{https://www.youtube.com/watch?v=ObhK8lUfIlc}, 2018.

\bibitem{denuvo}
Irdeto, ``Denuvo,'' \url{https://irdeto.com/denuvo/}, accessed February 2022.

\bibitem{csoverwatch}
``Counter strike: Global offensive. overwatch faq,''
  \url{https://blog.counter-strike.net/index.php/overwatch/}, accessed February
  2022.

\bibitem{duda2012pattern-chapter2}
R.~O. Duda, P.~E. Hart, and D.~G. Stork, \emph{Pattern classification}.\hskip
  1em plus 0.5em minus 0.4em\relax John Wiley \& Sons, 2012.

\bibitem{brummer2010measuring}
N.~Brummer, ``Measuring, refining and calibrating speaker and language
  information extracted from speech,'' Ph.D. dissertation, Stellenbosch:
  University of Stellenbosch, 2010.

\bibitem{ffavoidbans}
``Unknowncheats---avoid fairfight bans,''
  \url{https://unknowncheats.me/forum/infestation-survivor-stories/101840-iss-avoid-fairfight-bans-more.html},
  accessed February 2022.

\bibitem{oord2016wavenet}
A.~van~den Oord, S.~Dieleman, H.~Zen, K.~Simonyan, O.~Vinyals, A.~Graves,
  N.~Kalchbrenner, A.~Senior, and K.~Kavukcuoglu, ``Wavenet: A generative model
  for raw audio,'' in \emph{9th ISCA Speech Synthesis Workshop}, 2016, pp.
  125--125.

\bibitem{rossler2019faceforensics++}
A.~Rossler, D.~Cozzolino, L.~Verdoliva, C.~Riess, J.~Thies, and M.~Nie{\ss}ner,
  ``Faceforensics++: Learning to detect manipulated facial images,'' in
  \emph{Computer Vision and Pattern Recognition}, 2019, pp. 1--11.

\bibitem{mirza2014conditional}
M.~Mirza and S.~Osindero, ``Conditional generative adversarial nets,''
  \emph{arXiv preprint arXiv:1411.1784}, 2014.

\bibitem{arjovsky2017wasserstein}
M.~Arjovsky, S.~Chintala, and L.~Bottou, ``Wasserstein generative adversarial
  networks,'' in \emph{International Conference on Machine Learning}, 2017, pp.
  214--223.

\bibitem{adam}
D.~P. Kingma and J.~Ba, ``Adam: A method for stochastic optimization,'' in
  \emph{International Conference on Learning Representations}, 2015.

\bibitem{vizdoom}
M.~Kempka, M.~Wydmuch, G.~Runc, J.~Toczek, and W.~Ja{\'s}kowski, ``Vizdoom: A
  doom-based ai research platform for visual reinforcement learning,'' in
  \emph{Conference on Computational Intelligence and Games}, 2016, pp. 1--8.

\bibitem{clevert2015fast}
D.-A. Clevert, T.~Unterthiner, and S.~Hochreiter, ``Fast and accurate deep
  network learning by exponential linear units,'' in \emph{International
  Conference on Learning Representations}, 2016.

\bibitem{auto-sklearn}
M.~Feurer, A.~Klein, K.~Eggensperger, J.~Springenberg, M.~Blum, and F.~Hutter,
  ``Efficient and robust automated machine learning,'' in \emph{Advances in
  Neural Information Processing Systems 28}, 2015, pp. 2962--2970.

\bibitem{scikit-learn}
F.~Pedregosa, G.~Varoquaux, A.~Gramfort, V.~Michel, B.~Thirion, O.~Grisel,
  M.~Blondel, P.~Prettenhofer, R.~Weiss, V.~Dubourg, J.~Vanderplas, A.~Passos,
  D.~Cournapeau, M.~Brucher, M.~Perrot, and E.~Duchesnay, ``Scikit-learn:
  Machine learning in {P}ython,'' \emph{Journal of Machine Learning Research},
  vol.~12, pp. 2825--2830, 2011.

\bibitem{martin1997det}
A.~Martin, G.~Doddington, T.~Kamm, M.~Ordowski, and M.~Przybocki, ``The {DET}
  curve in assessment of detection task performance,'' National Inst of
  Standards and Technology Gaithersburg MD, Tech. Rep., 1997.

\bibitem{doddington2000nist}
G.~R. Doddington, M.~A. Przybocki, A.~F. Martin, and D.~A. Reynolds, ``The nist
  speaker recognition evaluation--overview, methodology, systems, results,
  perspective,'' \emph{Speech communication}, vol.~31, no. 2-3, pp. 225--254,
  2000.

\bibitem{steam2020}
``Steam---2020 year in review,''
  \url{https://steamcommunity.com/groups/steamworks/announcements/detail/2961646623386540827},
  accessed February 2022.

\bibitem{goodfellow2014explaining}
I.~Goodfellow, J.~Shlens, and C.~Szegedy, ``Explaining and harnessing
  adversarial examples,'' in \emph{International Conference on Learning
  Representations}, 2015.

\bibitem{von2004telling}
L.~Von~Ahn, M.~Blum, and J.~Langford, ``Telling humans and computers apart
  automatically,'' \emph{Communications of the ACM}, vol.~47, no.~2, pp.
  56--60, 2004.

\bibitem{zi2019end}
Y.~Zi, H.~Gao, Z.~Cheng, and Y.~Liu, ``An end-to-end attack on text captchas,''
  \emph{Transactions on Information Forensics and Security}, vol.~15, pp.
  753--766, 2019.

\bibitem{akrout2019hacking}
I.~Akrout, A.~Feriani, and M.~Akrout, ``Hacking {G}oogle {r}e{CAPTCHA} v3 using
  reinforcement learning,'' in \emph{Conference on Reinforcement Learning and
  Decision Making}, 2019.

\bibitem{vicencio2014effectiveness}
R.~Vicencio-Moreira, R.~L. Mandryk, C.~Gutwin, and S.~Bateman, ``The
  effectiveness (or lack thereof) of aim-assist techniques in first-person
  shooter games,'' in \emph{Conference on Human Factors in Computing Systems},
  2014, pp. 937--946.

\bibitem{philip2009botprize}
P.~Hingston, ``The 2k botprize,'' in \emph{IEEE Symposium on Computational
  Intelligence and Games}, 2009.

\bibitem{bernardi2017game}
M.~L. Bernardi, M.~Cimitile, F.~Martinelli, F.~Mercaldo, J.~Cardoso,
  L.~Maciaszek, M.~van Sinderen, and E.~Cabello, ``Game bot detection in online
  role player game through behavioural features.'' in \emph{ICSOFT}, 2017, pp.
  50--60.

\bibitem{explainable-cheating}
T.~Jianrong, X.~Yu, Z.~Shiwei, X.~Yuhong, L.~Jianshi, W.~Runze, and
  F.~Changjie, ``{XAI}-driven explainable multi-view game cheating detection,''
  in \emph{Conference on Games}, 2020.

\bibitem{cook2021social}
M.~Cook, ``The social responsibility of game ai,'' in \emph{Conference on
  Games}, 2021.

\bibitem{hagendorff2021forbidden}
T.~Hagendorff, ``Forbidden knowledge in machine learning reflections on the
  limits of research and publication,'' \emph{AI \& SOCIETY}, vol.~36, no.~3,
  pp. 767--781, 2021.

\end{thebibliography}

\vfill

\end{document}